\documentclass[sigconf]{acmart}
\usepackage{CJKutf8}
\usepackage{amsmath}
\usepackage{xspace}
\usepackage{graphicx}
\usepackage{algorithm}
\usepackage{algorithmic}
\usepackage{footnote}
\usepackage{makecell}
\usepackage{makecell}
\usepackage{booktabs}
\usepackage{multirow}
\usepackage{utfsym}
\usepackage{enumitem}
\usepackage{subfigure}
\usepackage{hyperref}
\usepackage{enumitem}
\usepackage{balance}

\AtBeginDocument{%
  }

\setcopyright{acmcopyright}

\copyrightyear{2023} 
\acmYear{2023} 
\setcopyright{acmlicensed}\acmConference[SIGIR '23]{Proceedings of the 46th International ACM SIGIR Conference on Research and Development in Information Retrieval}{July 23--27, 2023}{Taipei, Taiwan}
\acmBooktitle{Proceedings of the 46th International ACM SIGIR Conference on Research and Development in Information Retrieval (SIGIR '23), July 23--27, 2023, Taipei, Taiwan}
\acmPrice{15.00}
\acmDOI{10.1145/3539618.3591700}
\acmISBN{978-1-4503-9408-6/23/07}

\begin{document}

\title{Incorporating Structured Sentences with Time-enhanced BERT for Fully-inductive Temporal Relation Prediction}

\author{Zhongwu Chen}
\affiliation{%
  \institution{National University of Defense Technology}
  \city{Changsha}
  \country{China}
}
\email{chenzhongwu20@nudt.edu.cn}

\author{Chengjin Xu}
\authornotemark[1]
\affiliation{
  \institution{International Digital Economy Academy}
  \city{Shenzhen}
  \country{China}}
\email{xuchengjin@idea.edu.cn}

\author{Fenglong Su}
\authornotemark[1]
\affiliation{%
  \institution{National University of Defense Technology}
  \city{Changsha}
  \country{China}
}
\email{sufenglong18@nudt.edu.cn}

\author{Zhen Huang}
\authornote{Corresponding authors}
\affiliation{ 
  \institution{National University of Defense Technology}
  \city{Changsha}
  \country{China}}
  \email{huangzhen@nudt.edu.cn}

\author{Yong Dou}
\affiliation{ 
  \institution{National University of Defense Technology}
  \city{Changsha}
  \country{China}}
  \email{douyong@nudt.edu.cn}

%
\renewcommand{\shortauthors}{Zhongwu Chen, et al.}

\begin{abstract}

Temporal relation prediction in incomplete temporal knowledge graphs (TKGs) is a popular temporal knowledge graph completion (TKGC) problem in both transductive and inductive settings. Traditional embedding-based TKGC models (TKGE) rely on structured connections and can only handle a fixed set of entities, i.e., the transductive setting. In the inductive setting where test TKGs contain emerging entities, the latest methods are based on symbolic rules or pre-trained language models (PLMs).
However, they suffer from being inflexible and not time-specific, respectively.
In this work,  we extend the fully-inductive setting, where entities in the training and test sets are totally disjoint, into TKGs and take a further step towards a more flexible and time-sensitive temporal relation prediction approach SST-BERT\,\textendash\,incorporating \textbf{S}tructured \textbf{S}entences with \textbf{T}ime-enhanced \textbf{BERT}.
Our model can obtain the entity history and implicitly learn rules in the semantic space by encoding structured sentences, solving the problem of inflexibility. 
We propose to use a \emph{time} \emph{masking} MLM task to pre-train BERT in a corpus rich in temporal tokens specially generated for TKGs, enhancing the time sensitivity of SST-BERT. 
To compute the  probability of occurrence of a target quadruple, we aggregate all its structured sentences from both temporal and semantic perspectives into a score. Experiments on the transductive datasets and newly generated fully-inductive benchmarks show that SST-BERT successfully improves over state-of-the-art baselines.
\looseness=-1
\end{abstract}

\maketitle

\setlength{\parskip}{0.2cm plus4mm minus3mm}

\section{Introduction}
Temporal knowledge graphs (TKGs), as one of the most popular ways to store knowledge, are structural fact databases in the form of entities and relations between them over time~\cite{Fensel2020KnowledgeGM}. 
Many temporal knowledge-intensive tasks, such as information retrieval~\cite{IR}, question answering~\cite{ma-etal-2022-open} and entity alignment~\cite{xusu1,xusu2},
can benefit from TKGs. Recently, many temporal knowledge graph completion (TKGC) methods~\cite{wenzi1,Teru2020InductiveRP,tupian5,han2021explainable,ATiSE,wenzi2} focus on temporal relation prediction, a task of predicting missing links through reasoning over the observed facts in TKGs. In the real world, each fact has temporal constraints indicating that the fact holds true within a specific time period or at a certain time point. The typical formats of a fact are ($\emph{s}$,$\emph{r}$,$\emph{o}$,$\emph{t}\_\emph{begin}$,$\emph{t}\_\emph{end}$) and ($\emph{s}$,$\emph{r}$,$\emph{o}$,$\emph{t}$) in different datasets. Traditional temporal knowledge graph embedding (TKGE) representation learning approaches in TKGs embed entities, relations and time into low-dimensional vector spaces and measure the plausibility of quadruples via inputting their embeddings into a score function~\cite{9416312}. \looseness=-1

However, knowledge of TKGs is ever-changing and the temporal information makes TKGs highly dynamic. 
In the real world scenario, the emergence of new entities in the development process over time creates the need for temporal relation prediction in the fully-inductive setting, where entities in training TKGs and test TKGs are totally disjoint. 
Those traditional TKGE methods optimise the representation for a fixed predefined set of entities, i.e, the transductive setting, so they fail in the more realistic fully-inductive setting.
Some inductive relation prediction methods over static KGs~\cite{GraIL} have been proposed, but they can not learn time information in TKGs.
The rule-based method TLogic~\cite{TLogic} applies temporal logical rules and thus obtains the fully-inductive ability, but TLogic suffers from the limitation of the inflexibility of symbolic logic.
Due to the prior knowledge in PLMs, some PLM-based models achieve better results.
However, PLMs are usually pre-trained in large-scale corpora, so they are not adapted to particular domains such as temporal signals in TKGs~\cite{timemask}.  The texts in their corpora also have no explicit temporal indications, which makes it difficult for PLM-based models to handle temporally-scoped facts in TKGs.

 \looseness=-1

To tackle these problems, we propose SST-BERT, a model that incorporates \textbf{S}tructured \textbf{S}entences (constructed from relation paths and historical descriptions) with \textbf{T}ime-enhanced \textbf{BERT} by our designed \emph{time} \emph{masking} strategy to solve fully-inductive temporal relation prediction task. Table~\ref{compare} illustrates the detailed differences. Relation paths between target entities and historical descriptions of target entities make SST-BERT consider rich structural information and temporal knowledge while reasoning. The structured sentences enable SST-BERT to learn in semantic space, overcoming the symbolic restrictions in TLogic. Through our proposed \emph{time} \emph{masking} pre-training task in a corpus consisting of sentences rich in time tokens, SST-BERT is more time-sensitive to facts in TKGs.
\looseness=-1

First, in order to consider both structural and linguistic knowledge, we convert relation paths connecting the subject and object into natural language form as a part of the structured sentences and then utilize the language comprehension skill of PLMs to encode the relationships among facts, since LAMA~\cite{petroni-etal-2019-language} has shown that factual knowledge can be recovered well from PLMs, requiring no schema engineering.
Relation paths offer structural information and induce the implicit semantics hidden in the structured connections. 
Therefore, our model SST-BERT can capture temporal development patterns like rules and reason over TKGs in semantic space.
\looseness=-1

\looseness=-1
\begin{table*}[!ht]
\centering
\resizebox{.97\textwidth}{!}{
\begin{tabular}{c|l|l|r|r|r|r|r}
\toprule
\multirow{2}*[-1.2ex]{\shortstack{ \textbf{Method}}}
&\multirow{2}*[-1.2ex]{\shortstack{ \bf Transductive \\ \bf Setting}}
&\multirow{2}*[-1.2ex]{\shortstack{ \bf Fully-inductive \\ \bf Setting}}
&\multicolumn{3}{c|}{\textbf{Reasoning Evidence}}
&\multirow{2}*[-1.2ex]{\shortstack{ \bf Time\\ \bf Sensitivity}}
&\multirow{2}*[-1.2ex]{\shortstack{ \bf Explainable}}
\\
 \cline{4-6}
&&
&\makecell*[c]{\shortstack{\bf Relation \\ \bf Paths}}
&\makecell*[c]{\shortstack{\bf Historical\\ \bf Descriptions}}
&\makecell*[c]{\shortstack{\bf Prior\\ \bf Knowledge}}
&&
\\
\midrule
\midrule
\multicolumn{1}{c|}{TComplEx(TKGE)}
&\multicolumn{1}{c}{\usym{1F5F8}}&\multicolumn{1}{c|}{\usym{2717}}
&\multicolumn{1}{c}{\usym{2717}}&\multicolumn{1}{c}{\usym{2717}}&\multicolumn{1}{c|}{\usym{2717}}
&\multicolumn{1}{c|}{\usym{1F5F8}}
&\multicolumn{1}{c}{\usym{2717}}\\

\multicolumn{1}{c|}{KG-BERT(PLM-based)}
&\multicolumn{1}{c}{\usym{1F5F8}}&\multicolumn{1}{c|}{\usym{2717}}
&\multicolumn{1}{c}{\usym{2717}}&\multicolumn{1}{c}{\usym{2717}}&\multicolumn{1}{c|}{\usym{1F5F8}}
&\multicolumn{1}{c|}{\usym{2717}}
&\multicolumn{1}{c}{\usym{2717}}\\

\multicolumn{1}{c|}{BERTRL(PLM-based)}
&\multicolumn{1}{c}{\usym{1F5F8}}&\multicolumn{1}{c|}{\usym{1F5F8}}
&\multicolumn{1}{c}{\usym{1F5F8}}&\multicolumn{1}{c}{\usym{2717}}&\multicolumn{1}{c|}{\usym{1F5F8}}
&\multicolumn{1}{c|}{\usym{2717}}
&\multicolumn{1}{c}{\usym{1F5F8}}\\

\multicolumn{1}{c|}{TLogic(rule-based)}
&\multicolumn{1}{c}{\usym{1F5F8}}&\multicolumn{1}{c|}{\usym{1F5F8}}
&\multicolumn{1}{c}{\usym{1F5F8}}&\multicolumn{1}{c}{\usym{2717}}&\multicolumn{1}{c|}{\usym{2717}}
&\multicolumn{1}{c|}{\usym{1F5F8}}
&\multicolumn{1}{c}{\usym{1F5F8}}\\

\midrule
\multicolumn{1}{c|}{SST-BERT}
&\multicolumn{1}{c}{\usym{1F5F8}}&\multicolumn{1}{c|}{\usym{1F5F8}}
&\multicolumn{1}{c}{\usym{1F5F8}}&\multicolumn{1}{c}{\usym{1F5F8}}&\multicolumn{1}{c|}{\usym{1F5F8}}
&\multicolumn{1}{c|}{\usym{1F5F8}}
&\multicolumn{1}{c}{\usym{1F5F8}}\\
\bottomrule
\end{tabular}
}
\caption{
\label{compare}Comparison of our model SST-BERT with other algorithms in terms of their capability of temporal relation prediction in the transductive and fully-inductive settings; considering relation paths, historical descriptions and prior knowledge while reasoning (Reasoning Evidence); whether they can capture temporal changes (Time Sensitivity); and their explainability.
}
\vspace{-0.5cm}
\end{table*}

Secondly, to further enrich background knowledge for the subject and object in a target quadruple, we treat one single edge around the subject or object which is not included in the generated paths as background description texts.
Previous PLM-based knowledge graph completion methods, KG-BERT~\cite{KG-BERT} and StAR~\cite{StAR}, use the definitions or attributes of entities in external knowledge bases as supporting knowledge. But there exists plenty of information redundant and temporally irrelevant to the target quadruple. To overcome this limitation, our basic idea is to make full use of the more relevant and easily accessible history of entities inside TKGs, instead of relying on noisy external resources outside TKGs. Thus, the edges that happened before the target quadruple can serve as directly relevant information and can be converted into ideal historical description texts of entities. These easily accessible descriptions inside TKGs make up for the inadequate relation paths and are more targeted to the target quadruple than external knowledge bases.
Moreover, historical descriptions provide emerging entities with fully-inductive descriptions and integrate them into TKGs via PLMs.
As shown in Figure~\ref{mainfig}, we convert all the target quadruples in the training graph with their relation paths connecting the entities and  historical descriptions of the entities into structured sentences. \looseness=-1

Thirdly, we propose a $\emph{time}$ $\emph{masking}$ MLM pre-training task to enhance the time sensitivity of PLMs to changes in facts over time. 
Different from traditional open-domain corpora, we generate one domain-specific corpus rich in time tokens for each training TKG.
In our generated corpus, each sentence is associated with $\emph{2n}$ or $\emph{n}$ special time tokens $\left( \emph{n} \ge \emph{2} \right)$, which depends on the selected dataset.
Our proposed pre-training module not only forces PLMs to focus on the representations of special time tokens but also injects domain-specific knowledge into the parameters of PLMs.
In experiments, we use the popular pre-trained language model, BERT~\cite{Devlin2019BERTPO}, and refer to the pre-trained one as $\emph{TempBERT}$.
Compared with BERTRL~\cite{BERTRL}, our pre-trained $\emph{TempBERT}$ is more time-sensitive to the input structured sentences.
\looseness=-1

Finally, we leverage $\emph{TempBERT}$ to encode different parts of the structured sentences to represent entities and relations in the semantic space and then aggregate all the sentences to compute the  probability of occurrence of target quadruples.
Traditional TKGC datasets are set up for the transductive setting. To evaluate the ability of SST-BERT to deal with emerging entities, in Section~\ref{sec:Datasets}, we introduce a series of newly generated benchmarks for the temporal relation prediction task in the fully-inductive setting for the first time. The main contributions of this paper are summarized as follows:\looseness=-1

\begin{itemize}[leftmargin=*]
\setlength{\itemsep}{7pt }
\item[$\bullet$] New entities continually emerge in temporal knowledge graphs (TKGs) because of the highly dynamic of TKGs. To the best of our knowledge, this is the first attempt to explore the fully-inductive setting for the temporal relation prediction task. We reconstruct four new fully-inductive benchmarks for each selected dataset. 
\looseness=-1

\vspace{0.1cm}
\item[$\bullet$] We identify the shortcuts of current TKGE, rule-based and PLM-based baselines. 

Our corpora rich in time tokens are specially generated for TKGs. Therefore, the time-oriented MLM pre-training task, \emph{time} \emph{masking}, in these corpora and the structured sentences served as inputs of SST-BERT solve the problems of baselines.

\looseness=-1
\vspace{0.1cm}
\item[$\bullet$] We leverage relation paths and historical descriptions inside TKGs to recover prior knowledge stored in BERT and capture rule-like relation patterns in the semantic space, forming the capability of temporal relation prediction in both transductive and fully-inductive settings. The experiments verify the high performance and robustness of our model SST-BERT.\looseness=-1
\end{itemize}

\section{Related work}
\subsection{Transductive TKGC Models}
Most existing TKGC methods
are embedding-based and 
are extended from distance-based KGC models or semantic matching KGC models to a certain extent. 
Distance-based KGC models intensively
use translation-based scoring functions and
measure the distance between two entities. A typical example, 
TransE~\cite{TransE} ,
defines each relation as a translation 
from the subject to the object.
Semantic matching KGC models, such as ComplEx~\cite{Complex} and
DistMult~\cite{Yang2014EmbeddingEA}, calculate the semantic similarity of representations.
\looseness=-1

Following TransE, TTransE~\cite{TTransE} and HyTE~\cite{hyte} encode time in the entity-relation dimensional spaces with time embeddings and temporal hyperplanes.
TA-DistMult~\cite{TA-DistMult}, a temporal extension of DistMult, learns one core tensor for each timestamp based on Tucker decomposition.
Specifically, DE-SimplE~\cite{goel2020diachronic} incorporates time into diachronic entity embeddings and can deal with event-based TKG datasets with timestamps like [2014/12/15], such as ICEWS ~\cite{ICEWS}. But it has issues when facing datasets Wikidata~\cite{Wikidata} and YAGO~\cite{YAGO}, since the facts in them have a start time and an end time. In contrast, TeRo~\cite{tero} and TeLM~\cite{TeLM} model facts involving time intervals like [2003/\#\#/\#\#, 2005/\#\#/\#\#] and can be generalized to ICEWS, becoming state-of-the-art TKGE models in the embedding-based paradigm.
In our model, we can address both two situations.
Following ComplEx, TComplEx~\cite{TComplEx} extends the regularised CP decomposition to order 4 tensor. However, these TKGE models are naturally transductive, since they embed a fixed set of components while training and cannot be generalized to unseen entities after training. \looseness=-1

\subsection{Fully-inductive TKGC Models}
In contrast to the transductive setting, the inductive setting focuses on continually emerging new entities in TKGs. Graph Neural Network (GNN) is proven to be powerful in capturing non-linear architecture in KGs. 
However, they cannot handle  entirely new graphs, i.e., the fully-inductive setting, since there are no overlapping entities.
For the fully-inductive setting, GraIL~\cite{GraIL} extracts static KG subgraphs independent of any specific entities and trains a GNN as a score function.
Unlike GNN-based methods, the current rule-based TKGC approach TLogic~\cite{TLogic} induces probabilistic logic rules and applies learned node-independent rules to completely unseen entities. \looseness=-1

Another line of fully-inductive relation prediction is to introduce additional descriptions to embed unseen entities~\cite{DKRL,PKGC,dsecrip}. 
These methods use definitions or attributes of entities, which are imprecise for the target quadruples and highly costly to obtain.
In this paper, we utilize relation paths and historical descriptions in TKGs as texts and construct them into structured sentences for target entities and relations. In this way, we can easily recover their precise knowledge inside TKGs in the semantic space for the fully-inductive setting. \looseness=-1

\subsection{KG-enhanced Pre-trained Language Models}
Pre-trained language models (PLMs) on behalf of BERT~\cite{Devlin2019BERTPO} are trained in open-domain corpora and show effectiveness in capturing general language representations. But they lack domain-specific knowledge~\cite{KG-BERT}.
Recently, many works have investigated how to combine knowledge in KGs with PLMs. 
A popular approach to make PLMs more entity-aware is to introduce entity-aware objectives while pre-training. 
ERNIE~\cite{Sun2019ERNIE2A}, CokeBERT~\cite{SU2021} and CoLAKE~\cite{colake} predict the entity mentions to entities in texts with a cross-entropy loss or max-margin loss. 
KG-BERT~\cite{KG-BERT} and StAR~\cite{StAR} fine-tune  BERT  to incorporate information from the factual triples in KGs. SimKGC~\cite{simkgc} and C-LMKE~\cite{LMKE} leverage contrastive learning in a batch to model the probability that answers match questions.
However, none of them addressed the issue of lacking temporal knowledge in PLMs. \looseness=-1

\section{Method}

\subsection{Framework}

Temporal knowledge graphs (TKGs) consist of a set of edges {($\emph{s}$,\, $\emph{r}$, \, $\emph{o}$,\\
$\emph{t}\_\emph{begin}$, $\emph{t}\_\emph{end}$)} or ($\emph{s}$, $\emph{r}$, $\emph{o}$, $\emph{t}$) with
head, tail entities 
$\emph{s}$, $\emph{o}$ $\in$ $\mathcal{E}$ (the set of entities) and
relation 
$\emph{r}$ $\in$ $\mathcal{R}$ (the set of relations). Time period [$\emph{t}\_\emph{begin}$,$\emph{t}\_\emph{end}$] and timestamp $\emph{t}$ indicate when ($\emph{s}$,\, $\emph{r}$, \, $\emph{o}$) occurs because the fact may develop from time to time. 
The temporal relation prediction task in an incomplete temporal knowledge graph $\mathcal{G}$ is to score the probability that a missing edge 
called target quadruple ($\emph{t}\_\emph{begin}$ and $\emph{t}\_\emph{end}$ are treated as a whole) is true.\looseness=-1

\begin{figure*}[!htbp]
    \centering

    \resizebox{56em}{!}{\includegraphics{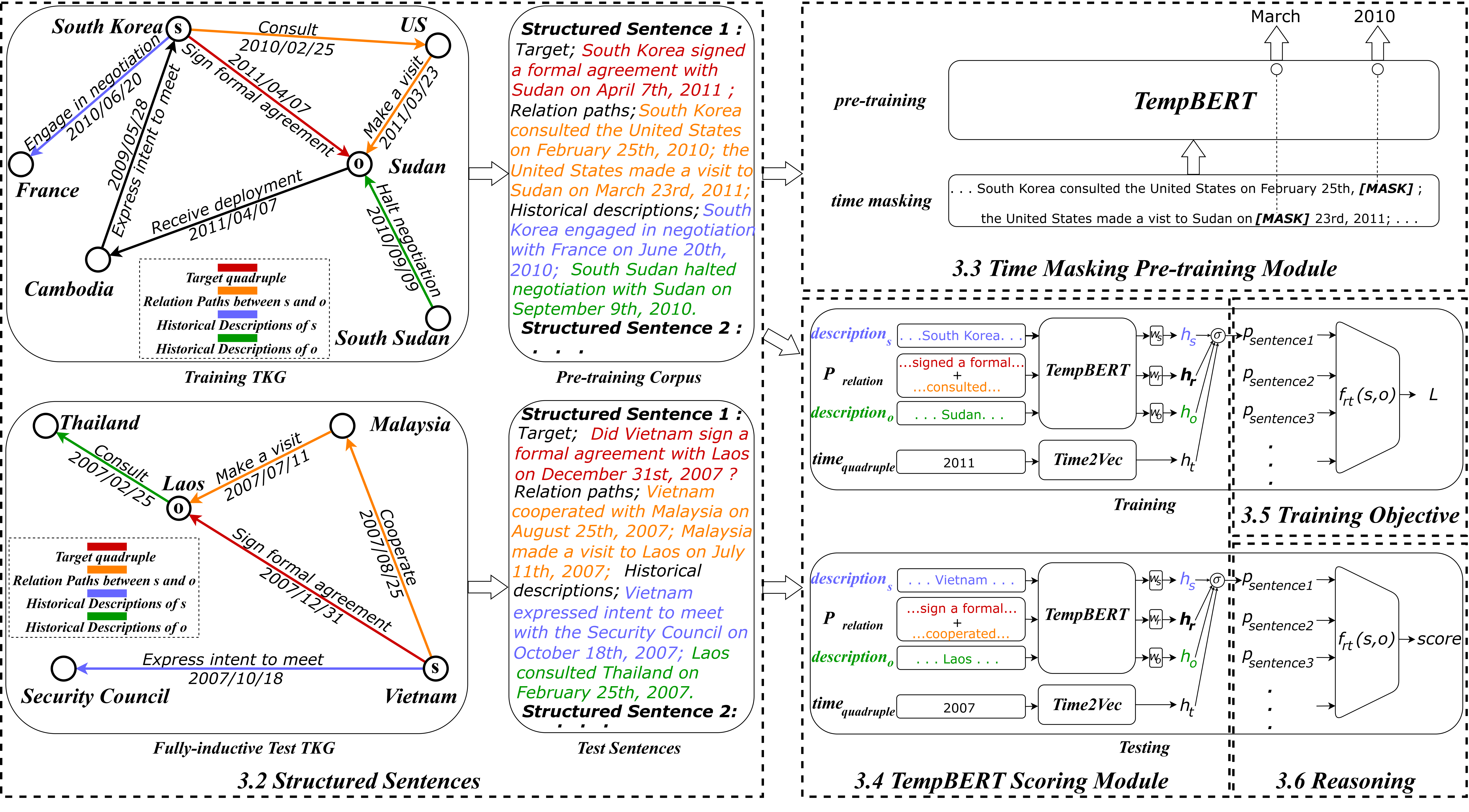}}
    \vspace{-0.1cm}
    \caption{Framework of our model SST-BERT. Each structured sentence consists of four parts: \textcolor[RGB]{202,0,0}{red} stands for the target quadruple ($\emph{s}$, $\emph{r}$, $\emph{o}$, $\emph{t}$); \textcolor[RGB]{255,128,0}{orange} stands for the relation paths between head entity $\emph{s}$ and tail entity $\emph{o}$; 
    \textcolor[RGB]{102,102,255}{blue} stands for the historical descriptions of head entity $\emph{s}$, \textcolor[RGB]{102,102,255}{$\emph{description}_\emph{s}$}; \textcolor[RGB]{0, 153, 0}{green} stands for the historical descriptions of tail entity $\emph{o}$, \textcolor[RGB]{0, 153, 0}{$\emph{description}_\emph{o}$}. 
    }
    \label{mainfig}
    \vspace{-0.2cm}
\end{figure*}

Our model scores a target quadruple 
in four steps as shown in Figure~\ref{mainfig}. 
First, we extract the relation paths between target entities $\emph{s}$ and $\emph{o}$ and historical descriptions of $\emph{s}$ and $\emph{o}$, denoted as G($\emph{s}$, $\emph{o}$) and then we convert them into structured sentences (Section~\ref{sec:Paths and Historical Knowledge Extracting}). Secondly, these structured sentences are used as a pre-training corpus to pre-train BERT by the proposed time-oriented MLM task, $\emph{time}$ $\emph{masking}$ (Section~\ref{sec:3.3}). The pre-trained BERT is called $\emph{TempBERT}$. Thirdly, we encode G($\emph{s}$, $\emph{o}$) with  $\emph{TempBERT}$ into the representations of entities and relations to compute the plausibility of each individual sentence.
The explicit times of the target quadruples are encoded by $\mathbf{t2v}$ (Section~\ref{sec:3.4}). Finally,  we design an aggregation function to combine all structured sentences related to a target quadruple for both training (loss) and testing (score). The score is the occurrence probability of the target quadruple (Section~\ref{sec:3.5},~\ref{sec:3.6}).
\looseness=-1

\subsection{Structured Sentences}
\label{sec:Paths and Historical Knowledge Extracting}

The structured sentences G($\emph{s}$, $\emph{o}$) surrounding entities $\emph{s}$ and $\emph{o}$ in a temporal knowledge graph  $\mathcal{G}$ provide important insights into predicting missing links between $\emph{s}$ and $\emph{o}$.
Traditional methods exploited a small scale of subgraphs to summarize relation patterns. For instance,  GraIL~\cite{GraIL}, CoMPILE~\cite{CoMPILE} and INDIGO~\cite{INDIGO} discussed subgraph-based relation reasoning and offered first-order logic. 
However, these subgraph-based approaches involve a paradox of ensuring the necessary patterns of relations are included while producing a sufficiently small graph. In practice, the presence of hub nodes often generates an accumulation of thousands of edges, which does not fit into the GPU memory for PLMs. Furthermore, we find that not all entity pairs have a subgraph, especially in the fully-inductive setting, leading to the failure of the subgraph-based methods.

Therefore, we adopt a more scalable and universal relation path-based reasoning approach to model the relationship between $\emph{s}$ and $\emph{o}$. 
Specifically, in Figure~\ref{mainfig} we convert symbolic edges into natural language
sentences
and capture the semantic logic hidden behind the relation paths.
In addition, motivated by the prompt-based models~\cite{autoprompt}, we manually define a prompt template for each relation type to express the semantics of quadruples more grammatically, rather than simply splicing their labels in previous works. 
Compared with symbolic rules explicitly mined by TLogic~\cite{TLogic}, our constructed structured sentences are more flexible to handle complex associations between facts in semantic space and are not formally restricted.\looseness=-1

In addition, temporal reasoning not only needs to consider what happened between target entities (relation paths), but should also pay attention to the history of target entities (historical descriptions).
In our PLM-based method, we argue that the old facts in the perspective of time offer vital background information for the target quadruples. 
Therefore, in Figure~\ref{mainfig} we randomly select one edge  which is not included in the relation paths as an old fact and regard it as a historical description to reflect the evolution of facts. 
Usually, these historical descriptions are semantically relevant to target quadruples and act as rich historical knowledge related to target entities. 
Furthermore, in the fully-inductive setting, the degrees of most entities are small, which means the ideal paths connecting target entities are limited and result in little information. Our proposed incorporation  of the history of entities enriches the range of information available to SST-BERT and makes interpretable inferences. We restrict their timestamps in the structured sentences to precede the timestamp(s) of the target quadruple to avoid information leakage.
As Figure~\ref{mainfig} shows, the structured sentences in training TKGs serve as both the pre-training corpus for \emph{time} \emph{masking} and the training inputs in the following sections.\looseness=-1

\subsection{Time Masking Pre-training Module}
\label{sec:3.3}

There is plenty of temporal information expressed in numerical form (day, year) or terminologies (month) in TKGs, which indicate the happening periods or timestamps of facts. 
They are represented in the temporal dimension in the embedding-based methods.
However, PLMs are pre-trained in noisy and generally purposed open-
\looseness=-1
domain corpora, so PLMs are not effective for applications on time-specific downstream tasks, such as temporal relation prediction.
In order to enhance the capability of PLMs on event temporal reasoning over TKGs, we creatively generate a small-scale pre-training corpus based on the training TKG and then pre-train PLMs in the corpus for the specific temporal domain to improve time sensitivity. 
\looseness=-1

Different from the pre-training corpus of the traditional pre-training task, our new corpus is constructed for the temporal domain in TKGs. 
We regard the whole training quadruple set as a big temporal graph and the relation paths and historical descriptions extracted for each edge correspond to a sentence in the new corpus. Section~\ref{sec:Paths and Historical Knowledge Extracting} provides the way to produce: we use bidirectional random walks to find paths and randomly select edges that are not included in the paths as historical descriptions of two target entities. Then we convert them into a natural language form and record the positions of timestamps.
We limit the relation paths and historical descriptions of two entities to three hops and one edge, respectively, to restrict the length of a sentence. Because each edge in the training TKGs is considered positive, all the generated sentences for it can be regarded as logically supporting the establishment of the edge to some extent. The timestamp information which is of key importance in our corpus can be easily obtained from all edges, so it becomes a frequent type of element and benefits our pre-training objective.\looseness=-1


Different from the masking strategy of the traditional pre-training task, we present a new MLM pre-training task, \emph{time} \emph{masking}, which replaces the random masking in the original MLM task with a time tokens targeted masking strategy. First, all positions of time tokens in each sentence $ \rho =\left(x_{1}, x_{2}, \ldots, x_{q}\right)$ are recorded in the corpus, so we can easily obtain the positions of temporal expressions denoted by $ T=\left(t_{1}, t_{2}, \ldots, t_{m}\right)$. Secondly, unlike in the case of BERT where 15\% of all tokens are randomly sampled, we 
focus on the time tokens, and 25\% of the temporal expressions in T are randomly sampled. We choose 25\% as we usually have more than four timestamps in the structured sentences, so we in most cases can mask at least one time token. Thirdly, we continuously randomly sample other tokens in $\rho-T$, until 15\% of all the tokens are sampled. Finally, we replace the sampled time tokens in $T$ with [MASK]; 80\% of the sampled tokens in $\rho-T$ are replaced with [MASK], 10\% with random tokens, and 10\% with the original tokens. Note that the sampled time tokens must be masked because recovering them helps PLMs to explore the time order of timestamps in the semantic space and identify the temporal relationships between events.

In practice, we use the popular PLM, BERT~\cite{Devlin2019BERTPO}, which we initialize from  $BERT_{BASE}$ (cased) in the publicly available website\footnote{\url{https://huggingface.co/}} for simplicity and efficiency. Our pre-trained BERT is called $\emph{TempBERT}$. 
Through our proposed \emph{time} \emph{masking} strategy, $\emph{TempBERT}$ is encouraged to pay attention to the explicit timestamps as well as the temporal changes of facts. The newly generated pre-training corpus rich in time information enables $\emph{TempBERT}$ to comprehend the evolution  of events in TKGs from the perspective of time, thus improving its time sensitivity. As a result, our time-enhanced BERT, $\emph{TempBERT}$, can be well adapted to the temporal relation prediction task. \looseness=-1

\subsection{TempBERT Scoring Module}
\label{sec:3.4}

In our model, we use $\emph{TempBERT}$ as an encoder to represent entities and relations.
As Figure~\ref{mainfig} shows, a structured sentence $\rho$ is divided into four parts. The texts converted from target quadruple and relation paths form the rule-like evolution of relation patterns, so they enhance the $\emph{TempBERT}$\,'s understanding of the target relation $r$. We combine token sequences of target quadruple and relation paths as  $ P_{relation} =([CLS], p^{r}_{1}, p^{r}_{2}, \ldots, p^{r}_{\ell}, [SEP]) $ and regard $ P_{relation}$ as the enhancement of $r$.
Description texts for head entity $\emph{s}$, $\emph{description}_{s} =([CLS], d^{s}_{1}, d^{s}_{2}, \ldots, d^{s}_{k},$ $[SEP]) $ and description texts for tail entity $\emph{o}$, $\emph{description}_{o} =([CLS],$ 
$ d^{o}_{1}, d^{o}_{2}, \ldots, d^{o}_{m}, [SEP]) $ are historical descriptions. They can also be considered as the enhancement of entities.
In BERT, the output embedding of token `[CLS]' aggregates information of the whole sequence, so we regard the three output embeddings of  token `[CLS]' in $\emph{description}_{s}$, $P_{relation}$ and $\emph{description}_{o}$  as embeddings $\mathbf{u}_{s}$, $\mathbf{u}_{r}$ and $\mathbf{u}_{o}$, respectively, to stand for the enhanced encoding by $\emph{TempBERT}$.
We use three independent linear layers to get head entity representation $\mathbf{h}_{s}$ $\in$ $\mathbb{R}^{d}$, relation representation $\mathbf{h}_{r}$ $\in$ $\mathbb{R}^{d}$ and  tail entity representation $\mathbf{h}_{o}$ $\in$ $\mathbb{R}^{d}$:\looseness=-1
\begin{equation*}
\begin{split}
\mathbf{h}_{s}&=\mathbf{w}_{s} \mathbf{u}_{s} + \mathbf{b}_{s};
\\
\mathbf{h}_{r}&=\mathbf{w}_{r} \mathbf{u}_{r} + \mathbf{b}_{r};
\\
\mathbf{h}_{o}&=\mathbf{w}_{o} \mathbf{u}_{o} + \mathbf{b}_{o},
\end{split}
\end{equation*}
where $\mathbf{w}_{s}$, $\mathbf{w}_{r}$, $\mathbf{w}_{o}$ $\in$ $\mathbb{R}^{d*d}$ and $\mathbf{b}_{s}$, $\mathbf{b}_{r}$, $\mathbf{b}_{o}$ $\in$ $\mathbb{R}^{d}$ are learnable weights and biases, $d$ is the  dimension of the output of token  `[CLS]' in $\emph{TempBERT}$. Because $\emph{TempBERT}$ has been enhanced by temporal information and each sentence includes rich timestamps, our obtained representations of entities and relations are time-sensitive.
In order to learn the time of the target quadruples ${time}_{quadruple}$ $t$ explicitly, we use $\mathbf{t2v}$ in Time2Vec~\cite{2019arXiv190705321M}  to encode $t$, denoted as $\mathbf{h}_{t}$ $\in$ $\mathbb{R}^{d}$:\looseness=-1
\begin{equation*}
\begin{split}
\mathbf{h}_{t}[i]= \mathbf{t2v} (t)[i] =\left\{\begin{array}{ll}
\omega_{i} t+\varphi_{i}, & \text { if } i=0 \\
sin \left(\omega_{i} t+\varphi_{i}\right), & \text { if } 1 \leq i \leq d-1
\end{array}\right.
\end{split}
 \end{equation*}
where $\mathbf{h}_{t}[i]$ is the $i^{th}$ element of $\mathbf{h}_{t}$; $\omega_{i}$s and $\varphi_{i}$s are learnable parameters.
Now, we compute the probability $\mathbf{p}_{sentence \rho}$ of the structured sentence $\rho $ leading to the establishment of the target quadruple:\looseness=-1
\begin{equation*}
\mathbf{p}_{sentence \rho} = \sigma \left( \mathbf{w}_{\theta}^\mathrm{T} (\mathbf{h}_{s} \ast  \mathbf{h}_{t} + \mathbf{h}_{r} - \mathbf{h}_{o} \ast  \mathbf{h}_{t}) + \mathbf{b}_{\theta} \right)  ,
 \end{equation*}
 where $\mathbf{w}_{\theta}$ $\in$ $\mathbb{R}^{d}$, $\mathbf{b}_{\theta}$ $\in$ $\mathbb{R}$ are learnable parameters, $\sigma$ is the sigmoid function. 
A target quadruple usually generates a set of sentences, and all of these sentences should be considered together to indicate the truth of the target quadruple while reasoning. Since we take an individual sentence as a knowledge extraction approach, $\mathbf{p}_{sentence \rho}$ scores for the individual sentence $\rho$. In the next section, we aggregate all the probabilities of structured sentences for the target quadruple, making the prediction comprehensively considered.\looseness=-1
\begin{table*}[t]
\centering
\resizebox{\linewidth}{!}{
\Huge
\begin{tabular}{ccccccc|cccc|cccc|cccc}
        \toprule
        & & &\multicolumn{4}{c|}{\Huge{ICEWS14}}&\multicolumn{4}{c|}{ICEWS05-15}&\multicolumn{4}{c|}{YAGO11k}&\multicolumn{4}{c}{Wikidata12k}
        \\
        \cmidrule(r){4-19}
        & &   & Entities & Relations & Time Tokens &\multicolumn{1}{c|}{Links}  & Entities & Relations & Time Tokens & \multicolumn{1}{c|}{Links} & Entities & Relations & Time Tokens & \multicolumn{1}{c|}{Links} & Entities & Relations & Time Tokens & \multicolumn{1}{c}{Links} \\
        \midrule
        \multirow{2}*[-0.2ex]{transductive}& & trans-train    & 6,869          & 230 & \multirow{2}*[-0.2ex]{365}          & 72,826  & 10,094          & 251 & \multirow{2}*[-0.2ex]{4,017 }         & 460,876  & 10,623          & 10 & \multirow{2}*[-0.2ex]{189 }          & 16,406  & 12,554          & 24 & \multirow{2}*[-0.2ex]{232 }           & 32,497        \\
        && trans-test   & 2,599          & 161  &           & 8,963    & 4,877          & 207 &           & 45,858  & 2,720          & 10 &          & 2,051  & 4,297          & 20 &         & 4,062      \\
        \midrule
        \multirow{8}*[-0.5ex]{fully inductive} & \multirow{2}*[-0.3ex]{v1} & ind-train    & 1,199          & 133   & \multirow{2}*[-0.2ex]{365}          & 6,170  & 1,926          & 173   & \multirow{2}*[-0.2ex]{4,017 }          & 23,830& 6,772          & 10   & \multirow{2}*[-0.2ex]{189 }           & 10,067 & 6,467          & 24   & \multirow{2}*[-0.2ex]{232 }           & 17,772    \\
        && ind-test   & 163          & 35    &           & 916 & 814          & 65   &           & 9,668& 712          & 10   &           & 1,298& 1,093          & 19   &       & 2,657     \\

        \cmidrule(r){2-19}
        & \multirow{2}*[-0.2ex]{v2} & ind-train    & 1,980          & 166   & \multirow{2}*[-0.2ex]{365}          & 14,694  & 3,064          & 231   & \multirow{2}*[-0.2ex]{4,017 }         & 57,415& 3,964          & 10   & \multirow{2}*[-0.2ex]{189 }            & 4,587 & 3,399          & 23   & \multirow{2}*[-0.2ex]{232 }         & 8,369    \\
        & &ind-test   & 353          & 57    &          & 2,657 & 1,858          & 108   &           & 20,432& 1,504          & 10   &         & 1,860& 1,509          & 22   &          & 3,812     \\
        \cmidrule(r){2-19}
        & \multirow{2}*[-0.2ex]{v3} & ind-train    & 3,891          & 213   & \multirow{2}*[-0.2ex]{365}          & 23,836 & 6,266          & 251   & \multirow{2}*[-0.2ex]{4,017 }         & 191,189& 2,455          & 10   & \multirow{2}*[-0.2ex]{189 }         & 3,890 & 2,345          & 23   &\multirow{2}*[-0.2ex]{232 }           & 5,258    \\
        && ind-test   & 917          & 102    &          & 4,884 & 2,549          & 196   &           & 38,477& 2,312          & 10   &          & 2,225& 2,400          & 23   &          & 4,240     \\
        \cmidrule(r){2-19}
        & \multirow{2}*[-0.2ex]{v4} & ind-train    & 1,262          & 147   & \multirow{2}*[-0.2ex]{365}           & 2,350  & 2,822          & 229   & \multirow{2}*[-0.2ex]{4,017 }         & 46,558& 1,738          & 10   & \multirow{2}*[-0.2ex]{189 }            & 1,914 & 1,562          & 24   & \multirow{2}*[-0.2ex]{232 }             & 3,542    \\
        && ind-test   & 2,492          & 135    &           & 7,260 & 4,107          & 207   &           & 84,855& 3,280          & 10   &           & 3,429& 3,735          & 24   &         & 6,155     \\
        \bottomrule
    \end{tabular}
}
\caption{Statistics of four transductive datasets, ICEWS14, ICEWS05-15, YAGO11k, Wikidata12k, and four created fully-inductive benchmarks for each of them. `trans-' represents the transductive setting and `ind-' represents the fully-inductive setting.}
\label{statistics}
\vspace{-0.5cm}
\end{table*}
\subsection{Training Objective}
\label{sec:3.5}
\vspace{-0.4cm}
While training, the training graph provides positive quadruples.
Self-adversarial negative sampling has been proven to be quite effective for TKGC~\cite{Sun2018RotatEKG}, which adopts negative sampling for efficient optimization.
We randomly sample $n$ negative quadruples for each positive quadruple by corrupting its head or tail entity and ensuring they do not exist in the training TKGs.
These sampled negative entities are restricted within common 3-hop neighbours of the entities in target quadruples.  Our self-adversarial loss is as follows:\looseness=-1
\begin{equation*}
\begin{split}
L=-\log \sigma\left(\gamma-f_{rt}(\mathbf{s},\mathbf{o})\right)
-\sum_{i=1}^{n} \frac{1}{n} \log \sigma(f_{rt}( \mathbf{s}_{i}^{\prime},\mathbf{o}_{i}^{\prime})-\gamma),
\end{split}
\end{equation*}
 where $\gamma$ is the margin, $\sigma$ is the sigmoid function, $n$ is the number of negative samples, $({s}_{i}^{\prime}, r, {o}_{i}^{\prime}, {t}_{begin}, {t}_{end})$ or $({s}_{i}^{\prime}, r, {o}_{i}^{\prime}, t)$ is the $i^{th}$ negative sample. Note that we generate sentences for both positive and negative quadruples, and $f_{rt}(\mathbf{s},\mathbf{o})$ is the aggregation function for generated structured sentences $\mathscr{D}(\mathbf{s},\mathbf{o})$ to score each quadruple. We further constrain the number of sentences produced by each quadruple, which is a hyperparameter $N$, to avoid redundant information, i.e., $ \lvert \mathscr{D}(\mathbf{s},\mathbf{o}) \rvert \leq N$. For datasets in the form of $(s, r, o, {t}_{begin}, {t}_{end})$:
 \begin{equation*}
\begin{split}
f_{rt}(\mathbf{s},\mathbf{o}) = \sum_{\rho \in \mathscr{D}(\mathbf{s},\mathbf{o})} \left ( \frac{exp( {t}_{\rho} - {t}_{begin})   }{ 2\sum_{\hat{\rho} \in \mathscr{D}(\mathbf{s},\mathbf{o})}exp( {t}_{\hat{\rho}}-  {t}_{begin})}+\right.
\\
\left.\frac{exp({t}_{\rho} - {t}_{end}  )}{ 2\sum_{\hat{\rho} \in \mathscr{D}(\mathbf{s},\mathbf{o})}exp(  
{t}_{\hat{\rho}} - {t}_{end} )}  \right ) \mathbf{p}_{sentence \rho} ,
\end{split}
\end{equation*}
where ${t}_{\rho}$ denotes the earliest timestamp in the sentence $\rho$ and ${t}_{\rho}$(${t}_{\hat{\rho}}$)\,$\leq$\,${t}_{begin}$\,$\leq$\,${t}_{end}$. For datasets in the form of $(s, r, o, t)$:
\begin{equation*}
\begin{split}
f_{rt}(\mathbf{s},\mathbf{o}) = \sum_{\rho \in \mathscr{D}(\mathbf{s},\mathbf{o})} \left ( \frac{exp( {t}_{\rho} - t)   }{ \sum_{\hat{\rho} \in \mathscr{D}(\mathbf{s},\mathbf{o})}exp( {t}_{\hat{\rho}}- t)}  \right ) \mathbf{p}_{sentence \rho} ,
\end{split}
\end{equation*}
where ${t}_{\rho}$ denotes the earliest timestamp in the sentence $\rho$ and ${t}_{\rho}$(${t}_{\hat{\rho}}$)\,$\leq$\,$t$. The exponential weighting favours sentences with timestamps that are closer to the timestamp(s) of the target quadruple, since they are more likely to be directly relevant to the prediction. 

\subsection{Reasoning}
\label{sec:3.6}
For a prediction question ($\emph{s}$,$\emph{r}$,?,$\emph{t}\_\emph{begin}$,$\emph{t}\_\emph{end}$) or ($\emph{s}$,$\emph{r}$,?,$\emph{t}$), we replace the asked tail entity with a set of candidate entities sampled from the whole entity set. Generally, the number of candidate entities can be varied from only a few to all, and it depends on the practical needs. $f_{rt}(\mathbf{s},\mathbf{o})$ is leveraged to score each candidate quadruple through its structured sentences. Then, the entity ranking first is chosen as the answer, and the sentence corresponding to the maximal summation factor in $f_{rt}(\mathbf{s},\mathbf{o})$ provides the most faithful explanation.\looseness=-1
\section{Experiment}
\vspace{-0.2cm}

To demonstrate the temporal relation prediction capability of our model SST-BERT, we compare SST-BERT with some state-of-the-art baselines in 
the transductive and  fully-inductive settings, i.e., Transductive Temporal Relation Prediction  and Fully-inductive Temporal Relation Prediction. Moreover, traditional TKGC datasets are constructed for the transductive setting, so we are supposed to offer variants derived from them for the fully-inductive setting. 
By structured sentences, SST-BERT can capture entities' full-inductive relation paths and historical descriptions inside TKGs. We want to explore the performance of SST-BERT compared with existing PLM-based models which rely on low-quality entity definitions or attributes linking to knowledge bases outside TKGs to offer fully-inductive ability. 
Another noteworthy point of SST-BERT is that we specially pre-train BERT on the generated TKG-aimed corpus and use \emph{time} \emph{masking} strategy to enhance the time sensitivity of the pre-trained \emph{TempBERT}. In contrast, existing PLM-based models are not sensitive to the essential temporal information in TKGs.\looseness=-1

 \subsection{Transductive Datasets and New Fully-inductive Benchmarks}
 \label{sec:Datasets}
 \vspace{-0.2cm}
Popular TKGC datasets include ICEWS14, ICEWS05-15, YAGO11k, Wikidata12k~\cite{hyte}, YAGO15k and Wikidata11k  ~\cite{TA-DistMult}. Time intervals in YAGO15k and Wikidata11k only contain either start dates or end dates, shaped like \emph{occurSince 2000} or \emph{occurUntill 2002}. However, since our model is good at capturing rich and complex temporal information, we prefer to use more challenging YAGO11k and Wikidata12k, where most of the time intervals contain both start dates and end dates, shaped like [2000/\#\#/\#\#, 2002/\#\#/\#\#] or [2000/01/01, 2002/01/01].  Similar to the setting in TeRo~\cite{tero}, we only deal with year-level information in YAGO11k and Wikidata12k and treat year timestamps as 189 and 232 special time tokens in BERT. This setting not only balances the numbers of quadruples in different time tokens but also makes the pre-training more targeted. 
ICEWS14 and ICEWS05-15 are subsets of the event-based database, ICEWS~\cite{ICEWS}, with political events in 2014 and 2005\,\textasciitilde\,2015. These two datasets are filtered by selecting the most frequently occurring entities in graphs. Their time annotations are all day-level, shaped like [2004/12/24]. The numbers of special time tokens in them are 365 and 4017.\looseness=-1

Our selected original datasets ICEWS14, ICEWS05-15, YAGO11k and Wikidata12k are only suitable for the transductive setting since the entities in the standard test splits are a subset of the entities in the training splits. 
We create new fully-inductive benchmarks for each dataset by sampling two disjoint subgraphs from its TKG as \emph{ind-train} and \emph{ind-test}.
\emph{ind-train} and \emph{ind-test} have a disjoint set of entities, and relations of \emph{ind-test} are entirely included in those  of \emph{ind-train}.
Specifically, we uniformly sampled several entities  to serve as roots and take several random walks to expand them as seen entities in \emph{ind-train}. Then all the quadruples including the selected seen entities create \emph{ind-train}. Finally, we remove these quadruples from the whole TKG and sample \emph{ind-test} using the same way as \emph{ind-train}. 
Similar to the evaluation setting of GraIL~\cite{GraIL}, four fully-inductive benchmarks of each dataset are generated with increasing test sizes and different ratios of seen entities in \emph{ind-train} and unseen entities in \emph{ind-test} by adjusting the length of random walks for robust evaluation.
In the fully-inductive setting, a model is trained on \emph{ind-train} and tested on \emph{ind-test}. The statistics of four transductive datasets and their sixteen fully-inductive benchmarks are listed in Table~\ref{statistics}.\looseness=-1
\begin{table*}[t]
\centering

\resizebox{0.90\linewidth}{!}{
\large
\begin{tabular}{|c|c|c|c|c|c|c|c|c|c|c|c|c|}
\specialrule{0.1em}{1pt}{1pt}
\multicolumn{1}{c}{}& \multicolumn{3}{c}{ICEWS$14$}&\multicolumn{3}{c}{ICEWS$05$-$15$}&\multicolumn{3}{c}{YAGO$11$k}&\multicolumn{3}{c}{Wikidata$12$k}
\\
\cmidrule(r){2-4}  \cmidrule(r){5-7} \cmidrule(r){8-10} \cmidrule(r){11-13}

\multicolumn{1}{c}{} & \multicolumn{1}{c}{MRR} & \multicolumn{1}{c}{Hits@$1$}& \multicolumn{1}{c}{Hits@$3$} & \multicolumn{1}{c}{MRR} & \multicolumn{1}{c}{Hits@$1$} & \multicolumn{1}{c}{Hits@$3$}& \multicolumn{1}{c}{MRR} & \multicolumn{1}{c}{Hits@$1$}& \multicolumn{1}{c}{Hits@$3$}
 & \multicolumn{1}{c}{MRR} & \multicolumn{1}{c}{Hits@$1$} & \multicolumn{1}{c}{Hits@$3$}
 \\
\specialrule{0.05em}{1pt}{1pt}
 \multicolumn{1}{c}{\text{HyTE}~\cite{hyte}} & \multicolumn{1}{c}{$0.297$} & \multicolumn{1}{c}{$10.8$} & \multicolumn{1}{c}{$41.6$}  & \multicolumn{1}{c}{$0.316$} & \multicolumn{1}{c}{$11.6$} & \multicolumn{1}{c}{$44.5$} &\multicolumn{1}{c}{$0.136$}  &\multicolumn{1}{c}{$3.3$}& \multicolumn{1}{c}{$-$} &\multicolumn{1}{c}{$0.253$}  &\multicolumn{1}{c}{$14.7$} &\multicolumn{1}{c}{$-$}
 \\
 \multicolumn{1}{c}{\text{TA-DistMult}~\cite{TA-DistMult}} & \multicolumn{1}{c}{$0.477$} & \multicolumn{1}{c}{$36.3$} & \multicolumn{1}{c}{$-$}  & \multicolumn{1}{c}{$0.474$} & \multicolumn{1}{c}{$34.6$} & \multicolumn{1}{c}{$-$} &\multicolumn{1}{c}{$0.155$}  &\multicolumn{1}{c}{$9.8$}& \multicolumn{1}{c}{$-$} &\multicolumn{1}{c}{$0.230$}  &\multicolumn{1}{c}{$13.0$} &\multicolumn{1}{c}{$-$}
 \\
 \multicolumn{1}{c}{\text{TComplEx}~\cite{TComplEx}} & \multicolumn{1}{c}{$0.610$} & \multicolumn{1}{c}{$53.0$} & \multicolumn{1}{c}{$66.0$}  & \multicolumn{1}{c}{$0.660$} & \multicolumn{1}{c}{$59.0$} & \multicolumn{1}{c}{$71.0$} &\multicolumn{1}{c}{$0.185$}  &\multicolumn{1}{c}{$12.7$}& \multicolumn{1}{c}{$18.3$} &\multicolumn{1}{c}{$0.331$}  &\multicolumn{1}{c}{$23.3$} &\multicolumn{1}{c}{$35.7$}
 \\
  \multicolumn{1}{c}{\text{TeRo}~\cite{tero}} & \multicolumn{1}{c}{$0.562$} & \multicolumn{1}{c}{$46.8$} & \multicolumn{1}{c}{$62.1$}  & \multicolumn{1}{c}{$0.586$} & \multicolumn{1}{c}{$46.9$} & \multicolumn{1}{c}{$66.8$} &\multicolumn{1}{c}{$0.187$}  &\multicolumn{1}{c}{$12.1$}& \multicolumn{1}{c}{$19.7$} & \multicolumn{1}{c}{$0.299$} &\multicolumn{1}{c}{$19.8$}  &\multicolumn{1}{c}{$32.9$}
  \\
\multicolumn{1}{c}{\text{TeLM}~\cite{TeLM}} & \multicolumn{1}{c}{$0.625$} & \multicolumn{1}{c}{$54.5$} & \multicolumn{1}{c}{$67.3$}  & \multicolumn{1}{c}{$0.678$} & \multicolumn{1}{c}{$59.9$} & \multicolumn{1}{c}{$72.8$} &\multicolumn{1}{c}{$0.191$}  &\multicolumn{1}{c}{$12.9$}& \multicolumn{1}{c}{$19.4$} & \multicolumn{1}{c}{$0.332$} &\multicolumn{1}{c}{$23.1$}  &\multicolumn{1}{c}{$36.0$}
 \\
 \multicolumn{1}{c}{{TLogic}~\cite{TLogic}} & \multicolumn{1}{c}{$0.430$} & \multicolumn{1}{c}{$33.6$} & \multicolumn{1}{c}{$48.3$}  & \multicolumn{1}{c}{$0.470$} & \multicolumn{1}{c}{$36.2$} & \multicolumn{1}{c}{$53.1$} &\multicolumn{1}{c}{$-$}  &\multicolumn{1}{c}{$-$} & \multicolumn{1}{c}{$-$} & \multicolumn{1}{c}{$-$} &\multicolumn{1}{c}{$-$}  &\multicolumn{1}{c}{$-$}
 \\
 \specialrule{0.05em}{1pt}{1pt}

\multicolumn{1}{c}{{KG-BERT}~\cite{KG-BERT}} & \multicolumn{1}{c}{$0.523$} & \multicolumn{1}{c}{$50.9$} & \multicolumn{1}{c}{$65.2$}  & \multicolumn{1}{c}{$0.539$} & \multicolumn{1}{c}{$52.0$} & \multicolumn{1}{c}{$63.2$} &\multicolumn{1}{c}{$0.462 $} &\multicolumn{1}{c}{$43.5$} & \multicolumn{1}{c}{$49.6$} & \multicolumn{1}{c}{$0.552$} &\multicolumn{1}{c}{$52.9$}  &\multicolumn{1}{c}{$59.6$}
 \\
\multicolumn{1}{c}{{StAR(Self-Adp)}~\cite{StAR}} & \multicolumn{1}{c}{$0.565$} & \multicolumn{1}{c}{$53.2$} & \multicolumn{1}{c}{$66.8$}  & \multicolumn{1}{c}{$0.564$} & \multicolumn{1}{c}{$53.1$} & \multicolumn{1}{c}{$65.6$} &\multicolumn{1}{c}{$0.496$} &\multicolumn{1}{c}{$48.9$} & \multicolumn{1}{c}{$52.4$} & \multicolumn{1}{c}{$0.593$} &\multicolumn{1}{c}{$54.6$}  &\multicolumn{1}{c}{$53.8$}
 \\
 \multicolumn{1}{c}{{C-LMKE}~\cite{LMKE}} & \multicolumn{1}{c}{$0.576$} & \multicolumn{1}{c}{$47.0$} & \multicolumn{1}{c}{$65.8$}  & \multicolumn{1}{c}{$0.659$} & \multicolumn{1}{c}{$63.2$} & \multicolumn{1}{c}{$68.8$}&\multicolumn{1}{c}{$0.497$}  &\multicolumn{1}{c}{$46.3$}& \multicolumn{1}{c}{$50.1$} & \multicolumn{1}{c}{$0.648$} &\multicolumn{1}{c}{$60.0$}  &\multicolumn{1}{c}{$66.8$}
 \\
\multicolumn{1}{c}{BERTRL~\cite{BERTRL}} & \multicolumn{1}{c}{$0.635$} & \multicolumn{1}{c}{$60.9$} & \multicolumn{1}{c}{$68.9$}  & \multicolumn{1}{c}{$0.648$} & \multicolumn{1}{c}{$62.4$} & \multicolumn{1}{c}{$65.3$} &\multicolumn{1}{c}{$0.513$} &\multicolumn{1}{c}{$47.1$} & \multicolumn{1}{c}{$57.9$} & \multicolumn{1}{c}{$0.573$} &\multicolumn{1}{c}{$55.8$}  &\multicolumn{1}{c}{$58.6$}
 \\
\multicolumn{1}{c}{SimKGC~\cite{simkgc}} & \multicolumn{1}{c}{$0.396$} & \multicolumn{1}{c}{$31.4$} & \multicolumn{1}{c}{$42.9$}  & \multicolumn{1}{c}{$0.605$} & \multicolumn{1}{c}{$50.7$} & \multicolumn{1}{c}{$66.3$} &\multicolumn{1}{c}{$0.196$} &\multicolumn{1}{c}{$10.7$} & \multicolumn{1}{c}{$20.5$} & \multicolumn{1}{c}{$0.389$} &\multicolumn{1}{c}{$29.3$}  &\multicolumn{1}{c}{$42.4$}
 \\

\specialrule{0.05em}{1pt}{1pt}

\multicolumn{1}{c}{SST-BERT} & \multicolumn{1}{c}{\textbf{$0.688$}} & \multicolumn{1}{c}{\textbf{$62.4$}} & \multicolumn{1}{c}{\textbf{$72.1$}}  & \multicolumn{1}{c}{\textbf{$0.693$}} & \multicolumn{1}{c}{\textbf{$66.3$}} & \multicolumn{1}{c}{\textbf{$76.9$}} &\multicolumn{1}{c}{\textbf{$0.558$}}  &\multicolumn{1}{c}{\textbf{$50.5$}}& \multicolumn{1}{c}{\textbf{$58.2$}} & \multicolumn{1}{c}{\textbf{$0.684$}} &\multicolumn{1}{c}{\textbf{$65.9$}}  &\multicolumn{1}{c}{\textbf{$69.5$}}
\\
\specialrule{0.1em}{1pt}{1pt}

\end{tabular}}

\caption{\label{table1}Temporal relation prediction results (\% for Hits@$k$) on four datasets in the transductive setting. Dashes: results are not reported in the corresponding  works. Other results are obtained from our experiments. 
\textbf{Bold} numbers denote the best results. }
\vspace{-0.7cm}
\end{table*}
 \subsection{Baselines}
 \vspace{-0.2cm}
We compare SST-BERT with typical TKGE models,  HyTE~\cite{hyte}, TA-DistMult~\cite{TA-DistMult}, TComplEx~\cite{TComplEx}, TeRo~\cite{tero} and TeLM~\cite{TeLM}, for these TKGE models are the state-of-the-art methods in the transductive setting on our four selected datasets. In addition, these TKGE models embed time information into the temporal dimension from different views, becoming a traditional and popular paradigm. 
However, TKGE models are trained and tested within a fixed TKG with no new entities, so they can not be  used directly in the fully-inductive setting.
SST-BERT is also compared with a state-of-the-art symbolic framework for TKGC, TLogic~\cite{TLogic}, which is based on temporal logical rules extracted via temporal random walks.  Different from TKGE models, the learned rules in TLogic are entity-independent, making TLogic a fully-inductive method.
The results of TLogic on YAGO11k and Wikidata12k can not be obtained, since TLogic focuses on facts with one timestamp rather than facts in YAGO11k and Wikidata12k with start dates and end dates.
In SST-BERT, the pre-trained \emph{TempBERT} encodes the structured sentences consisting of relation paths and historical descriptions in semantic space to implicitly form rules, instead of in restricted symbolic space. 
\looseness=-1
\begin{table*}[t]
\centering

\tiny
\resizebox{0.90\linewidth}{!}{
\Large
\begin{tabular}{|c|c|c|c|c|c|c|c|c|c|c|c|c|c|c|c|c|}
\specialrule{0.1em}{1pt}{1pt}
\multicolumn{1}{c}{}& \multicolumn{4}{c}{ICEWS$14$}&\multicolumn{4}{c}{ICEWS$05$-$15$}&\multicolumn{4}{c}{YAGO$11$k}&\multicolumn{4}{c}{Wikidata$12$k}
\\
\cmidrule(r){2-5}  \cmidrule(r){6-9} \cmidrule(r){10-13} \cmidrule(r){14-17}

\multicolumn{1}{c}{} & \multicolumn{1}{c}{v$1$} & \multicolumn{1}{c}{v$2$}& \multicolumn{1}{c}{v$3$} & \multicolumn{1}{c}{v$4$} & \multicolumn{1}{c}{v$1$} & \multicolumn{1}{c}{v$2$}& \multicolumn{1}{c}{v$3$} & \multicolumn{1}{c}{v$4$} & \multicolumn{1}{c}{v$1$} & \multicolumn{1}{c}{v$2$}& \multicolumn{1}{c}{v$3$} & \multicolumn{1}{c}{v$4$} & \multicolumn{1}{c}{v$1$} & \multicolumn{1}{c}{v$2$}& \multicolumn{1}{c}{v$3$} & \multicolumn{1}{c}{v$4$}
 \\
\specialrule{0.05em}{1pt}{1pt}

 \multicolumn{1}{c}{TLogic~\cite{TLogic}} & \multicolumn{1}{c}{$0.153$} & \multicolumn{1}{c}{$0.173$} & \multicolumn{1}{c}{$0.228$}  & \multicolumn{1}{c}{$0.284$} & \multicolumn{1}{c}{$0.268$} & \multicolumn{1}{c}{$0.283$} &\multicolumn{1}{c}{$0.346$} &\multicolumn{1}{c}{$0.392$} & \multicolumn{1}{c}{$-$} & \multicolumn{1}{c}{$-$} &\multicolumn{1}{c}{$-$}  &\multicolumn{1}{c}{$-$}& \multicolumn{1}{c}{$-$} & \multicolumn{1}{c}{$-$} & \multicolumn{1}{c}{$-$}  & \multicolumn{1}{c}{$-$}
 \\
 \specialrule{0.05em}{1pt}{1pt}

\multicolumn{1}{c}{KG-BERT~\cite{KG-BERT}} & \multicolumn{1}{c}{$0.461$} & \multicolumn{1}{c}{$0.437$} & \multicolumn{1}{c}{$0.454$}  & \multicolumn{1}{c}{$0.422$} & \multicolumn{1}{c}{$0.565$} & \multicolumn{1}{c}{$0.536$} &\multicolumn{1}{c}{$0.594$} &\multicolumn{1}{c}{$0.538$} & \multicolumn{1}{c}{$0.465$} & \multicolumn{1}{c}{$0.495$} &\multicolumn{1}{c}{$0.435$}  &\multicolumn{1}{c}{$0.423$}& \multicolumn{1}{c}{$0.399$} & \multicolumn{1}{c}{$0.485$} & \multicolumn{1}{c}{$0.468$}  & \multicolumn{1}{c}{$0.342$}
 \\
\multicolumn{1}{c}{StAR(Self-Adp)~\cite{StAR}} & \multicolumn{1}{c}{$0.498$} & \multicolumn{1}{c}{$0.453$} & \multicolumn{1}{c}{$0.481$}  & \multicolumn{1}{c}{$0.464$} & \multicolumn{1}{c}{$0.592$} & \multicolumn{1}{c}{$0.562$} &\multicolumn{1}{c}{$0.584$} &\multicolumn{1}{c}{$0.524$} & \multicolumn{1}{c}{$0.482$} & \multicolumn{1}{c}{$0.462$} &\multicolumn{1}{c}{$0.521$}  &\multicolumn{1}{c}{$0.487$}& \multicolumn{1}{c}{$0.389$} & \multicolumn{1}{c}{$0.431$} & \multicolumn{1}{c}{$0.494$}  & \multicolumn{1}{c}{$0.405$}
 \\
 \multicolumn{1}{c}{C-LMKE~\cite{LMKE}} & \multicolumn{1}{c}{$0.523$} & \multicolumn{1}{c}{$0.542$} & \multicolumn{1}{c}{$0.582$}  & \multicolumn{1}{c}{$0.535$} & \multicolumn{1}{c}{$0.724$} & \multicolumn{1}{c}{$0.777$} &\multicolumn{1}{c}{$0.751$}  &\multicolumn{1}{c}{$0.706$}& \multicolumn{1}{c}{$0.535$} & \multicolumn{1}{c}{$0.552$} &\multicolumn{1}{c}{$0.527$}  &\multicolumn{1}{c}{$0.518$}& \multicolumn{1}{c}{$0.517$} & \multicolumn{1}{c}{$0.596$} & \multicolumn{1}{c}{$0.624$}  & \multicolumn{1}{c}{$0.546$}
 \\
\multicolumn{1}{c}{BERTRL~\cite{BERTRL}} & \multicolumn{1}{c}{$0.342$} & \multicolumn{1}{c}{$0.292$} & \multicolumn{1}{c}{$0.256$}  & \multicolumn{1}{c}{$0.237$} & \multicolumn{1}{c}{$0.685$} & \multicolumn{1}{c}{$0.640$} &\multicolumn{1}{c}{$0.623$} &\multicolumn{1}{c}{$0.652$} & \multicolumn{1}{c}{$0.530$} & \multicolumn{1}{c}{$0.556$} &\multicolumn{1}{c}{$0.549$}  &\multicolumn{1}{c}{$0.512$}& \multicolumn{1}{c}{$0.532$} & \multicolumn{1}{c}{$0.545$} & \multicolumn{1}{c}{$0.569$}  & \multicolumn{1}{c}{$0.467$}
 \\
\multicolumn{1}{c}{SimKGC~\cite{simkgc}} & \multicolumn{1}{c}{$0.064$} & \multicolumn{1}{c}{$0.028$} & \multicolumn{1}{c}{$0.030$}  & \multicolumn{1}{c}{$0.039$} & \multicolumn{1}{c}{$0.073$} & \multicolumn{1}{c}{$0.033$} &\multicolumn{1}{c}{$0.046$} &\multicolumn{1}{c}{$0.021$} & \multicolumn{1}{c}{$0.070$} & \multicolumn{1}{c}{$0.068$} &\multicolumn{1}{c}{$0.064$}  &\multicolumn{1}{c}{$0.059$}& \multicolumn{1}{c}{$0.052$} & \multicolumn{1}{c}{$0.081$} & \multicolumn{1}{c}{$0.074$}  & \multicolumn{1}{c}{$0.083$}
 \\

\specialrule{0.05em}{1pt}{1pt}

\multicolumn{1}{c}{$SST-BERT$} & \multicolumn{1}{c}{\textbf{$0.569$}} & \multicolumn{1}{c}{\textbf{$0.588$}} & \multicolumn{1}{c}{\textbf{$0.622$}}  & \multicolumn{1}{c}{\textbf{$0.613$}} & \multicolumn{1}{c}{\textbf{$0.732$}} & \multicolumn{1}{c}{\textbf{$0.798$}} &\multicolumn{1}{c}{\textbf{$0.765$}}  &\multicolumn{1}{c}{\textbf{$0.739$}}& \multicolumn{1}{c}{\textbf{$0.547$}} & \multicolumn{1}{c}{\textbf{$0.599$}} &\multicolumn{1}{c}{\textbf{$0.562$}}  &\multicolumn{1}{c}{\textbf{$0.584$}}& \multicolumn{1}{c}{\textbf{$0.551$}} & \multicolumn{1}{c}{\textbf{$0.619$}} & \multicolumn{1}{c}{\textbf{$0.644$}}  & \multicolumn{1}{c}{\textbf{$0.571$}}
\\
\specialrule{0.1em}{1pt}{1pt}

\end{tabular}}

\caption{\label{table2}Temporal relation prediction MRR results on the fully-inductive benchmarks. \textbf{Bold} numbers denote the best results.}
\vspace{-0.5cm}
\end{table*}

\begin{table*}[t]
\centering

\resizebox{0.90\linewidth}{!}{
\small 
\begin{tabular}{|c|c|c|c|c|c|c|c|c|c|c|c|c|c|c|c|c|}
\specialrule{0.1em}{1pt}{1pt}
\multicolumn{1}{c}{}& \multicolumn{4}{c}{ICEWS$14$}&\multicolumn{4}{c}{ICEWS$05$-$15$}&\multicolumn{4}{c}{YAGO$11$k}&\multicolumn{4}{c}{Wikidata$12$k}
\\
\cmidrule(r){2-5}  \cmidrule(r){6-9} \cmidrule(r){10-13} \cmidrule(r){14-17}

\multicolumn{1}{c}{} & \multicolumn{1}{c}{v$1$} & \multicolumn{1}{c}{v$2$}& \multicolumn{1}{c}{v$3$} & \multicolumn{1}{c}{v$4$} & \multicolumn{1}{c}{v$1$} & \multicolumn{1}{c}{v$2$}& \multicolumn{1}{c}{v$3$} & \multicolumn{1}{c}{v$4$} & \multicolumn{1}{c}{v$1$} & \multicolumn{1}{c}{v$2$}& \multicolumn{1}{c}{v$3$} & \multicolumn{1}{c}{v$4$} & \multicolumn{1}{c}{v$1$} & \multicolumn{1}{c}{v$2$}& \multicolumn{1}{c}{v$3$} & \multicolumn{1}{c}{v$4$}
 \\
\specialrule{0.05em}{1pt}{1pt}

 \multicolumn{1}{c}{TLogic~\cite{TLogic}} & \multicolumn{1}{c}{$13.2$} & \multicolumn{1}{c}{$15.3$} & \multicolumn{1}{c}{$19.5$}  & \multicolumn{1}{c}{$$26.6$$} & \multicolumn{1}{c}{$23.5$} & \multicolumn{1}{c}{$27.5$} &\multicolumn{1}{c}{$30.9$} &\multicolumn{1}{c}{$35.5$} & \multicolumn{1}{c}{$-$} & \multicolumn{1}{c}{$-$} &\multicolumn{1}{c}{$-$}  &\multicolumn{1}{c}{$-$}& \multicolumn{1}{c}{$-$} & \multicolumn{1}{c}{$-$} & \multicolumn{1}{c}{$-$}  & \multicolumn{1}{c}{$-$}
 \\
 \specialrule{0.05em}{1pt}{1pt}

\multicolumn{1}{c}{KG-BERT~\cite{KG-BERT}} & \multicolumn{1}{c}{$45.9$} & \multicolumn{1}{c}{$42.5$} & \multicolumn{1}{c}{$43.7$}  & \multicolumn{1}{c}{$40.1$} & \multicolumn{1}{c}{$52.9$} & \multicolumn{1}{c}{$52.6$} &\multicolumn{1}{c}{$55.7$} &\multicolumn{1}{c}{$50.4$} & \multicolumn{1}{c}{$42.8$} & \multicolumn{1}{c}{$46.2$} &\multicolumn{1}{c}{$40.1$}  &\multicolumn{1}{c}{$38.9$}& \multicolumn{1}{c}{$32.8$} & \multicolumn{1}{c}{$41.2$} & \multicolumn{1}{c}{$42.5$}  & \multicolumn{1}{c}{$30.1$}
 \\
\multicolumn{1}{c}{StAR(Self-Adp)~\cite{StAR}} & \multicolumn{1}{c}{$45.2$} & \multicolumn{1}{c}{$41.6$} & \multicolumn{1}{c}{$45.3$}  & \multicolumn{1}{c}{$42.9$} & \multicolumn{1}{c}{$52.4$} & \multicolumn{1}{c}{$51.6$} &\multicolumn{1}{c}{$54.9$} &\multicolumn{1}{c}{$50.9$} & \multicolumn{1}{c}{$45.2$} & \multicolumn{1}{c}{$49.3$} &\multicolumn{1}{c}{$50.4$}  &\multicolumn{1}{c}{$43.7$}& \multicolumn{1}{c}{$34.6$} & \multicolumn{1}{c}{$35.4$} & \multicolumn{1}{c}{$35.6$}  & \multicolumn{1}{c}{$38.6$}
 \\
 \multicolumn{1}{c}{C-LMKE~\cite{LMKE}} & \multicolumn{1}{c}{$46.9$} & \multicolumn{1}{c}{$53.8$} & \multicolumn{1}{c}{$56.7$}  & \multicolumn{1}{c}{$46.2$} & \multicolumn{1}{c}{$62.3$} & \multicolumn{1}{c}{$63.7$} &\multicolumn{1}{c}{$62.2$}  &\multicolumn{1}{c}{$58.5$}& \multicolumn{1}{c}{$49.3$} & \multicolumn{1}{c}{$51.1$} &\multicolumn{1}{c}{$47.5$}  &\multicolumn{1}{c}{$44.9$}& \multicolumn{1}{c}{$47.0$} & \multicolumn{1}{c}{$53.9$} & \multicolumn{1}{c}{$51.7$}  & \multicolumn{1}{c}{$54.2$}
 \\
\multicolumn{1}{c}{BERTRL~\cite{BERTRL}} & \multicolumn{1}{c}{$22.9$} & \multicolumn{1}{c}{$20.2$} & \multicolumn{1}{c}{$24.5$}  & \multicolumn{1}{c}{$15.4$} & \multicolumn{1}{c}{$61.2$} & \multicolumn{1}{c}{$53.4$} &\multicolumn{1}{c}{$60.3$} &\multicolumn{1}{c}{$62.9$} & \multicolumn{1}{c}{$50.2$} & \multicolumn{1}{c}{$57.9$} &\multicolumn{1}{c}{$52.0$}  &\multicolumn{1}{c}{$49.6$}& \multicolumn{1}{c}{$47.5$} & \multicolumn{1}{c}{$51.1$} & \multicolumn{1}{c}{$43.2$}  & \multicolumn{1}{c}{$42.6$}
 \\
\multicolumn{1}{c}{SimKGC~\cite{simkgc}} & \multicolumn{1}{c}{$2.2$} & \multicolumn{1}{c}{$0.6$} & \multicolumn{1}{c}{$0.4$}  & \multicolumn{1}{c}{$0.5$} & \multicolumn{1}{c}{$4.1$} & \multicolumn{1}{c}{$1.4$} &\multicolumn{1}{c}{$2.7$} &\multicolumn{1}{c}{$0.4$} & \multicolumn{1}{c}{$2.2$} & \multicolumn{1}{c}{$2.4$} &\multicolumn{1}{c}{$1.8$} &\multicolumn{1}{c}{$1.9$}& \multicolumn{1}{c}{$2.0$} & \multicolumn{1}{c}{$3.1$} & \multicolumn{1}{c}{$2.8$}  & \multicolumn{1}{c}{$3.2$}
 \\

\specialrule{0.05em}{1pt}{1pt}

\multicolumn{1}{c}{SST-BERT} & \multicolumn{1}{c}{\textbf{$54.4$}} & \multicolumn{1}{c}{\textbf{$56.9$}} & \multicolumn{1}{c}{\textbf{$60.8$}}  & \multicolumn{1}{c}{\textbf{$59.6$}} & \multicolumn{1}{c}{\textbf{$67.1$}} & \multicolumn{1}{c}{\textbf{$66.9$}} &\multicolumn{1}{c}{\textbf{$65.9$}}  &\multicolumn{1}{c}{\textbf{$63.2$}}& \multicolumn{1}{c}{\textbf{$53.2$}} & \multicolumn{1}{c}{\textbf{$59.1$}} &\multicolumn{1}{c}{\textbf{$54.9$}}  &\multicolumn{1}{c}{\textbf{$52.6$}}& \multicolumn{1}{c}{\textbf{$51.1$}} & \multicolumn{1}{c}{\textbf{$57.2$}} & \multicolumn{1}{c}{\textbf{$53.4$}}  & \multicolumn{1}{c}{\textbf{$56.9$}}
\\
\specialrule{0.1em}{1pt}{1pt}

\end{tabular}}

\caption{\label{table3}Temporal relation prediction Hits@$1$ results on the fully-inductive benchmarks. \textbf{Bold} numbers denote the best results.}
\vspace{-0.8cm}
\end{table*}

Since our model SST-BERT 
brings BERT and TKG together,
we compare it with some current PLM-based models. These approaches leverage BERT to encode entities using description texts, so they are naturally inductive. Some PLM-based TKGC models, such as KG-BRET~\cite{KG-BERT}, StAR~\cite{StAR} and BERTRL~\cite{BERTRL}, input external supplementary descriptions or path information into PLMs. Others, such as SimKGC~\cite{simkgc} and C-LMKE~\cite{LMKE}, combine PLMs with contrastive learning. We thus choose them as baselines. Since PLMs in these baselines are designed to encode all kinds of information, different from their original static datasets, we adapt them to our temporal datasets. The external texts of entities required by KG-BRET, StAR and SimKGC are retrieved from Wikipedia\footnote{\url{https://en.wikipedia.org/}}, because entity definitions or attributes outside TKGs are a core part of the three models.\looseness=-1

\subsection{Implementation Details}
 \vspace{-0.1cm}
 For each dataset, we generate a small-scale but domain-specific corpus following Section~\ref{sec:Paths and Historical Knowledge Extracting} and pre-train $BERT_{BASE}$ (cased) with 12 layers and 110M parameters on it for 10 epochs using our proposed $\emph{time}$ $\emph{masking}$ strategy to enhance the time sensitivity of BERT. 
 The maximum sequence length of the input tokens is 512. The batch size is 8. We use AdamW~\cite{Kingma2014AdamAM} as the optimizer and set the learning rate to be $5{e}^{-4}$, with gradient accumulation equal to 32. The above pre-training phase was carried out on 32G Tesla V100 GPUs.\looseness=-1

After pre-training, the obtained $\emph{TempBERT}$ is utilized as an encoder to embed entities and relations for scoring. In the training phase, the batch size is chosen from $\left\{16, 32, 64\right\}$  and the learning rate is selected from $\left\{ 2{e}^{-4}, 1{e}^{-4}, 5{e}^{-5}, 1{e}^{-5} \right\}$. The number of negative quadruples for training $n$ is tuned in $\left\{5, 10, 20, 50\right\}$, and the maximum number of sentences produced by each quadruple $N$ is in a range
of $\left\{10, 20, 50, 100\right\}$. The output dimension $d$ of $\emph{TempBERT}$ and $\mathbf{t2v}$ is 768. \looseness=-1

\subsection{Experimental Setup}
\vspace{-0.1cm}
 PLM-based methods use BERT as an encoder throughout the process of running. In order to speed up the evaluation process, we generate 50 negative samples for each quadruple in the validation set in advance. These negative samples are generated by corrupting head or tail entities in each quadruple and then selecting other entities in 3-hop neighbourhoods.
 While testing, we select all entities as candidates. The final results are obtained by ranking the ground truth among the negative quadruples by decreasing scores.
 We compute the mean reciprocal rank (MRR) and Hits@$k$ for $k$ $\in$ $\left\{1, 3 \right\}$.  \looseness=-1

\subsection{Transductive Temporal Relation Prediction}
In this section, we explore the performance of our model SST-BERT in the transductive setting in Table~\ref{table1}.
TKGE models embed time into vector space  and are originally designed for the transductive setting. They form  powerful baselines in the temporal relation prediction task. Compared with TKGE models, SST-BERT considers time and easily accessible knowledge inside TKGs in the semantic space and outperforms them by a stable margin.
TLogic~\cite{TLogic} is a current rule-based model which utilizes entity-independent rules. Our generated relation paths and historical descriptions  deal with rule-like reasoning more flexibly and the semantic understanding capability makes the results of SST-BERT higher than TLogic. 
Compared with PLM-based baselines which directly use BERT as an encoder without adaptation, SST-BERT enhances itself with time sensitivity by pre-training on the corpus generated from the target dataset TKGs and achieves considerable performance gains over PLM-based baselines.\looseness=-1

\subsection{Fully-inductive Temporal Relation Prediction}
The fully-inductive setting requires models not only to understand the logical patterns of relations but also to be capable of generalizing to unseen entities.
Our proposed framework captures the intrinsic evolutionary patterns of relations in the semantic space via the pre-trained $\emph{TempBERT}$ and encodes the emerging entities over time through generated structured sentences. Therefore, our model
can be naturally applied to the fully-inductive setting.
In reality, since we can not control the number or the ratio of newly emerging entities, we conduct experiments on the four generated  benchmarks of each dataset  with increasing test sizes and different ratios of seen entities and unseen entities  to verify the robustness of our model. \looseness=-1

From Table~\ref{table2} and Table~\ref{table3}, we can find our model SST-BERT outperforms all baselines in the fully-inductive setting.
In TKGE models, new entities can not be sufficiently trained, so we exclude them.
TLogic~\cite{TLogic} relies on matching the data in the test set with the found rules, so the larger the amount of data in the test set, the more likely TLogic is to give more correct answers.
However,  the emerging entities are usually much fewer than the total entities in the transductive setting and the edges linking to the emerging entities are also limited. Therefore, compared with the transductive setting,  the performance of TLogic decreases in the fully-inductive setting. 
Previous PLM-based models, KG-BERT~\cite{KG-BERT}, StAR~\cite{StAR} and C-LMKE~\cite{LMKE}, focus on triplet-level information and heavily rely on co-reference resolution and entity linking. They require external data (texts from Wikipedia), causing the accumulation of errors~\cite{nguyen}. 
Although BERTRL~\cite{BERTRL} considers paths, 
the important historical descriptions for temporal relation prediction are still ignored. 
In-batch Negatives (IB) module is a key part of SimKGC$_{\mathrm{IB}+\mathrm{PB}+\mathrm{SN}}$~\cite{simkgc}, but this module is disabled when handling emerging entities. This means not all PLM-based models are suitable for the fully-inductive setting.\looseness=-1

As the experiments show, our model SST-BERT has a strong fully-inductive ability to reason flexibly over complicated TKGs based on implicitly learned  semantic rules.
We combine the structured sentences and prior knowledge in PLMs, and the overall framework is similar in the transductive  and fully-inductive  settings, so there is no bias in the temporal relation prediction capability of SST-BERT in the two settings. 
In addition, SST-BERT has the most robust results among all baselines except SimKGC$_{\mathrm{IB}+\mathrm{PB}+\mathrm{SN}}$ due to its low performance. For example, the variances of the Hit@$1$ results of SST-BERT are 6.14, 2.42, 6.45 and 6.43 for the four datasets. In contrast, all the variances of BERTRL are greater than 10.
Therefore, SST-BERT can be robustly adapted to the fully-inductive setting in the real world.\looseness=-1

\vspace{-0.3cm}
\subsection{Ablation study}
\subsubsection{Effect of relation paths, historical descriptions and t2v}
\ 
\newline
As mentioned in Section~\ref{sec:Paths and Historical Knowledge Extracting}, 
our generated structured sentences consist of relation paths and historical descriptions, these converted texts in the natural language form provide both structural and semantic knowledge. 
Moreover, relation paths serve as the source of fully-inductive ability and historical descriptions additionally offer background information.
To emphasize the significance of the two key components of our model SST-BERT, we conduct an ablation study to remove relation paths (w/o paths) and remove historical descriptions (w/o  history). 
By removing $\mathbf{t2v}$ (w/o $\mathbf{t2v}$), we also illustrate to what extent the explicitly encoded time contributes to SST-BERT.
The results~\ref{Ablation} show that removing any part will reduce the performance.
Two kinds of descriptions are vital for SST-BERT to recall the time-oriented knowledge stored in the PLMs, proving that temporal descriptions of entities are remarkable for the time-sensitive TKGC task. 
The role of $\mathbf{t2v}$ is to make SST-BERT precisely recognize the time of the target quadruples without being distracted by other factors in the structured sentences.
\looseness=-1

\subsubsection{Effect of time masking pre-training task}
\label{sec:Effect of time masking pre-training}
\begin{figure}[t]
	\centering
  \vspace{-0.2cm}
        \vspace{-0.1cm}

        \begin{minipage}{0.49\linewidth}
		\centering
            \hspace{-5.5cm plus30mm} 
             \includegraphics[scale=0.01]{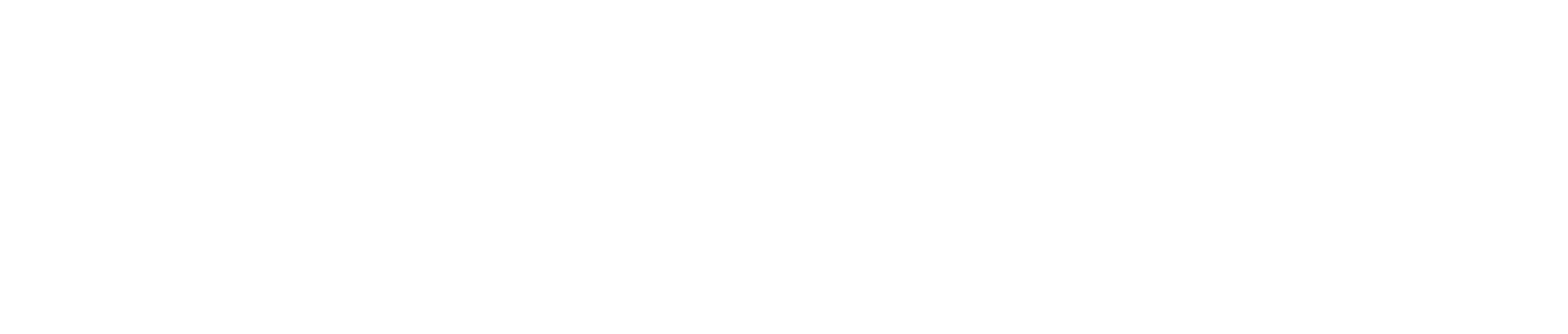}

		\label{chutian2}
	\end{minipage}
	\begin{minipage}{\linewidth}
            \hspace{0.35cm} 

		\includegraphics[scale=0.3]{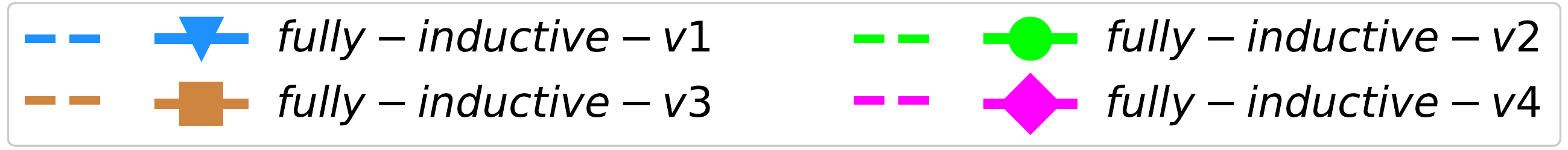}

		\label{chutian1}
	\end{minipage}

 \vspace{0.3cm}
        \hspace{-0.03cm }

	\begin{minipage}{0.49\linewidth}
		\centering
             \vspace{-0.2cm}

            \includegraphics[scale=0.12]{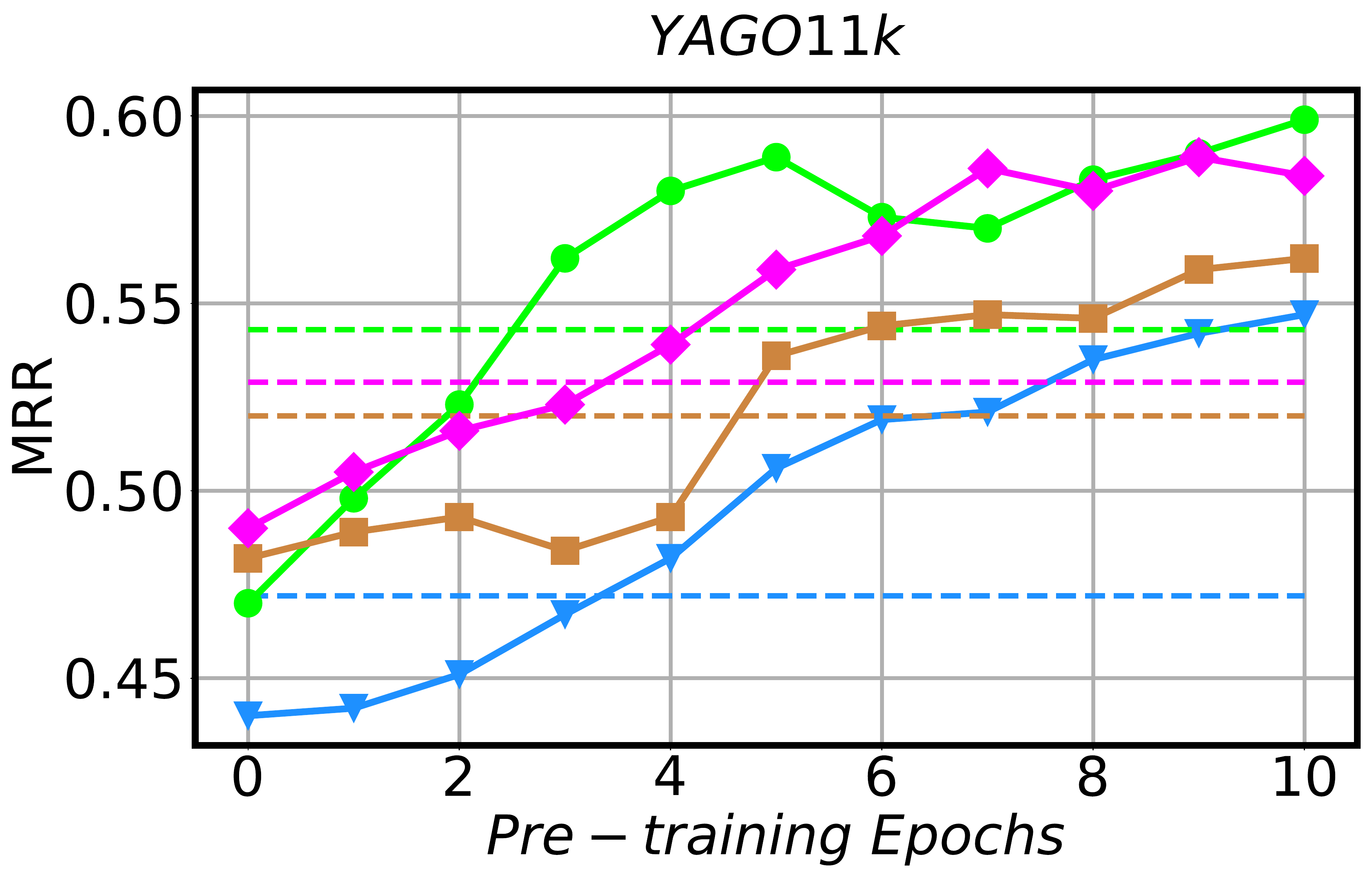}
            
            \hspace{0.3cm} 
		\label{chutian3}
	\end{minipage}
	\begin{minipage}{0.49\linewidth}
		\centering
            \vspace{-0.2cm}
            \hspace{-0.05cm} 	
            
            \includegraphics[scale=0.12]{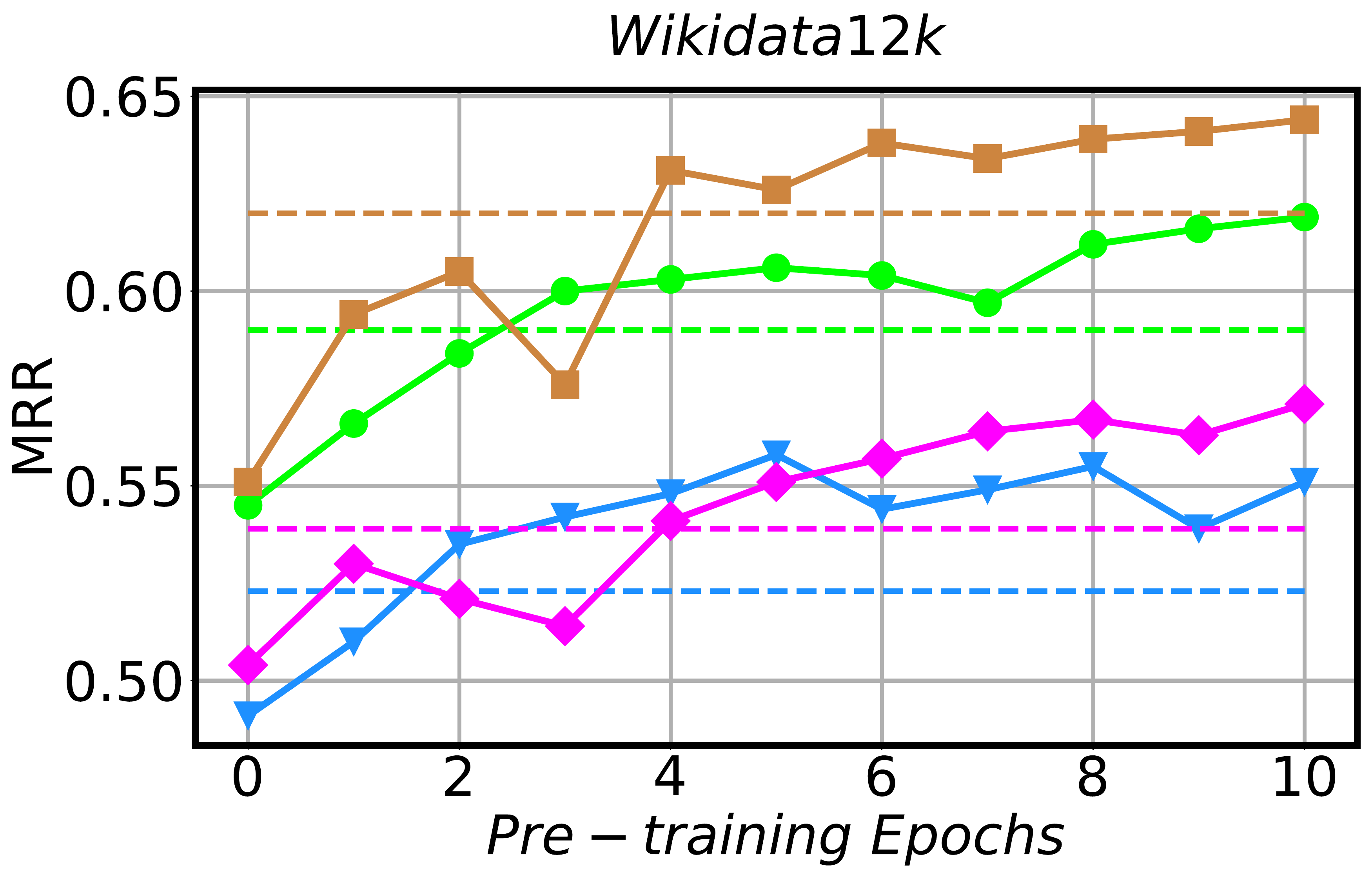}
		\label{chutian4}
	\end{minipage}
 \vspace{-0.7cm}
     \caption{\label{line}The dashed line represents the best MRR results  of SST-BERT within 10 pre-training epochs with the original random masking pre-training task in BERT\cite{Devlin2019BERTPO}. The solid line represents the MRR results of SST-BERT with our proposed \emph{time} \emph{masking} pre-training task from 0 to 10 epochs. The same colour represents two results are under the same benchmark.\looseness=-1}
     \vspace{-0.0cm}
\end{figure}

\ 
\newline
In this section, we explore the performance improvements from the pre-trained $\emph{TempBERT}$  with the growth of pre-training epochs. In Figure~\ref{line}, we replace the \emph{time} \emph{masking} pre-training task with the original random masking pre-training task in BERT~\cite{Devlin2019BERTPO} to illustrate our contribution. 
First, both pre-training tasks improve the performance of SST-BERT compared with no pre-training. Furthermore, our proposed \emph{time} \emph{masking} outperforms the best results gotten by original random masking after 2\,\textasciitilde\,5 epochs due to the time-type targeted masking.
Secondly, SST-BERT observes obvious growth in the early pre-training epochs, and the final performance gains for the two datasets are about 21.9\% and 14.0\% on average.
Therefore, our proposed \emph{time} \emph{masking} is efficient and effective.
In our experiments, we choose to pre-train 10 epochs, since we find the performance tends to be relatively stable after pre-training 10 epochs and the cost of pre-training and the performance achieved are well balanced.\looseness=-1

\subsection{Explainability and Case Study}
\vspace{-0.1cm}
Temporal logical rules are usually considered to be explainable. TLogic~\cite{TLogic} utilizes temporal random walks to get temporal logical rules in symbolic form.
When TLogic applies the found rules to answer questions, it must precisely match each relation and timestamp in the rules, including relation types, relation order and time order. If there are no matching body groundings in the graph, then no answers will be predicted for the given questions. 
In contrast, SST-BERT leverages generated relation paths and historical descriptions to form rules in the semantic space, achieving explainability. While pre-training, our proposed $\emph{time}$ $\emph{masking}$ MLM task increases the ability of BERT to understand the meaning of timestamps. Furthermore, through training, we implicitly inject semantic rules into the parameters, which are more flexible than symbolic rules.
As long as the semantics of the two edges are similar, SST-BERT can detect the relevance of rules rather than abandon them.
Finally, for a prediction task $(s, r, ?, {t}_{begin}, {t}_{end})$ or $(s, r, ?, t)$, we aggregate all the sentences found and their probabilities  for different candidate tail entities. The tail entity $o$ with the highest probability is chosen as the answer and the relations paths and historical descriptions are regarded as the explanation of the occurrence  of the target quadruple.\looseness=-1

\begin{table}[t]
\centering
\resizebox{\linewidth}{!}{
\begin{tabular}{ccccccc}
\specialrule{0.1em}{1pt}{1pt}
fully-inductive benchmarks& SST-BERT & w/o paths & w/o history & w/o t2v\\
\specialrule{0.05em}{1pt}{1pt}
ICEWS14 (v2)& 0.588 & 0.536  & 0.491  & 0.552 \\
ICEWS05-15 (v2)& 0.798 & 0.724 & 0.680 & 0.743 \\
\specialrule{0.1em}{1pt}{1pt}
\end{tabular}}
\caption{\label{Ablation}Ablation study of our model SST-BERT (MRR).}
\vspace{-0.6cm}
\end{table}

\begin{figure}[!]
    \centering
\vspace{-0.1cm}
    \includegraphics
    [width=\linewidth]
    {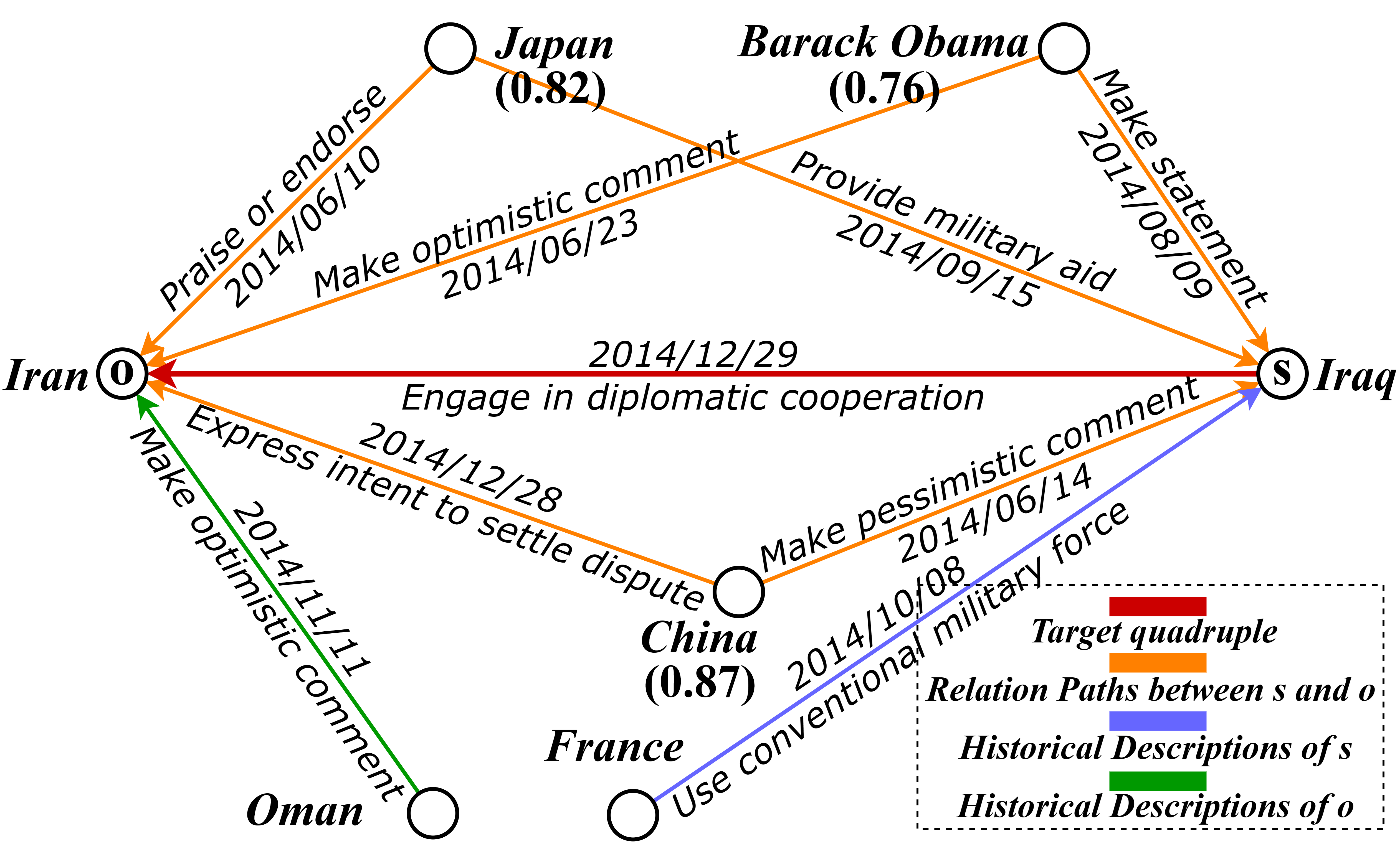}
    \caption{Explainability for a prediction case (\emph{Iraq}, \emph{Engage} \emph{in} \emph{diplomatic} \emph{cooperation}, ?, \emph{2014-12-29}) and its answer \emph{Iran}. Numbers in brackets stand for the probabilities $\mathbf{p}_{sentence \rho}$s of the structured sentences generated by SST-BERT.}
    \label{case}
    \vspace{-0cm}
\end{figure}

For a prediction question (\emph{Iraq}, \emph{Engage} \emph{in} \emph{diplomatic} \emph{cooperation}, ?, \emph{2014/12/29}), Figure~\ref{case} illustrates the reasoning evidence for the ground truth answer \emph{Iran}. We can see that there are two countries, \emph{China} and \emph{Japan}, and one person, \emph{Barack} \emph{Obama}, engaged in the happening of the target quadruple; there are two countries, \emph{France} and \emph{Oman}, offering historical descriptions for the target entities. 
First, the three relation paths and the two selected historical descriptions construct three structured sentences for the target quadruple, which have the probabilities $\mathbf{p}_{sentence \rho}$s of 0.87, 0.82, 0.76, ranking top 3. The final score given by $f_{rt}(\mathbf{s},\mathbf{o})$ is 0.83, ranking first among candidate entities.
Secondly, two historical facts, (\emph{Japan}, \emph{Provide} \emph{military} \emph{aid}, \emph{Iraq}, \emph{2014/9/15}) and (\emph{France}, \emph{Use} \emph{conventional} \emph{military} \emph{force}, \emph{Iraq}, \emph{2014/10/08}), show the whole affair was military related. Finally, the edge (\emph{China}, \emph{Express} \emph{intent} \emph{to} \emph{settle} \emph{dispute}, \emph{Iran}, \emph{2014/12/28})  temporally and semantically  closest to the target quadruple was the direct cause of the occurrence of
\emph{engaging} \emph{in} \emph{diplomatic} \emph{cooperation} between \emph{Iraq} and \emph{Iran}.\looseness=-1


\section{Conclusion}
In this paper, our model SST-BERT incorporates structured sentences with time-enhanced BERT as a comprehensively considered and time-sensitive solution to predict missing temporal relations and extends to the fully-inductive setting over TKGs.
The rule-like relation paths in the natural language form enable SST-BERT to reason in a flexible and explainable way. 
The historical descriptions enhance the target entities with easily accessible historical information inside TKGs and make SST-BERT external resource-independent.
\emph{TempBERT} pre-trained  by our proposed \emph{time} \emph{masking} strategy in a specially generated TKG-aimed 
corpus rich in time tokens makes SST-BERT more sensitive to the time changing
than BERT used in PLM-based baselines. 
We generate various benchmarks of the four datasets for the fully-inductive setting to evaluate SST-BERT comprehensively.
Experiments show the outperformance of SST-BERT and the effectiveness of our proposed modules.
Moreover, SST-BERT is robust to the different test sizes and different ratios of seen and unseen entities in the fully-inductive setting.\looseness=-1

\bibliographystyle{ACM-Reference-Format}
\balance
\bibliography{mybib}


\end{document}